\documentclass[runningheads]{llncs}

\usepackage[T1]{fontenc}
\usepackage{graphicx}
\usepackage{amsmath}
\usepackage{amssymb}
\usepackage{booktabs}
\usepackage{subcaption}
\usepackage{float}
\usepackage{colortbl}
\usepackage[table,xcdraw]{xcolor}
\usepackage{caption}
\usepackage{hyperref}

\definecolor{babyblue}{rgb}{0.54, 0.81, 0.94}

\begin{document}

\title{LITE: A Paradigm Shift in Multi-Object Tracking with Efficient ReID Feature Integration}

\author{Jumabek Alikhanov\inst{1}\orcidID{0000-0003-3103-6033}, Dilshod Obidov\inst{1}\orcidID{0009-0007-9845-6979} \and
Hakil Kim\inst{1}\orcidID{0000-0003-4232-3804}}
\authorrunning{J. Alikhanov et al.}
\institute{Department of Electrical and Computer Engineering, Inha University, Incheon, 22212 South Korea\\
This research was supported and funded by HUMBLEBEE R\&D.\\
\email{juma@inha.edu, dilshod@humblebee.ai, hikim@inha.ac.kr}\\
\textit{Corresponding author: Hakil Kim (e-mail: hikim@inha.ac.kr)}
}

\maketitle              

\begin{abstract}
The Lightweight Integrated Tracking-Feature Extraction (LITE) paradigm is introduced as a novel multi-object tracking (MOT) approach. It enhances ReID-based trackers by eliminating inference, pre-processing, post-processing, and ReID model training costs. LITE uses real-time appearance features without compromising speed. By integrating appearance feature extraction directly into the tracking pipeline using standard CNN-based detectors such as YOLOv8m, LITE demonstrates significant performance improvements. The simplest implementation of LITE on top of classic DeepSORT achieves a HOTA score of 43.03\% at 28.3 FPS on the MOT17 benchmark, making it twice as fast as DeepSORT on MOT17 and four times faster on the more crowded MOT20 dataset, while maintaining similar accuracy. Additionally, a new evaluation framework for tracking-by-detection approaches reveals that conventional trackers like DeepSORT remain competitive with modern state-of-the-art trackers when evaluated under fair conditions. The code will be available post-publication at \url{https://github.com/Jumabek/LITE}.

\end{abstract}


\section{Introduction}
Multiple Object Tracking (MOT) is crucial in computer vision for maintaining consistent object identities across video frames, with applications in surveillance, autonomous driving, and sports analytics \cite{ciaparrone2020deep,luo2021multiple}. Real-time performance is essential in these scenarios to ensure timely and accurate responses to dynamic events \cite{meva_challenge,yu2022argus++}.
Furthermore, strong MOT systems and can advances the field of Assisted Living technologies and deep learning enhanced elderly care. MOT enables reliablly monitoring for fall detection and vital signs \cite{CLIMENTPEREZ2020112847,sathyanarayana2018vision}.

Re-Identification (ReID) aims to associate objects, particularly humans, across different camera views. It is vital in scenarios where the same object may leave the field of view of one camera and reappear in another, such as in multi-camera surveillance systems, tracking athletes in sports, and analyzing customer behavior in retail environments. The challenge lies in matching identities accurately despite variations in lighting, pose, occlusions, and camera angles. Current ReID-based trackers face issues like unclear improvement contributions, dependency on advanced detectors, and lack of exhaustive evaluation under varied conditions \cite{zhang2022bytetrack,du2023strongsort}. Although recent methods \cite{zhang2022bytetrack,maggiolino2023deep} show superior performance, they often require specific detector training and sophisticated loss functions.

\begin{figure*}[h]
\centering
\fbox{\includegraphics[width=\linewidth]{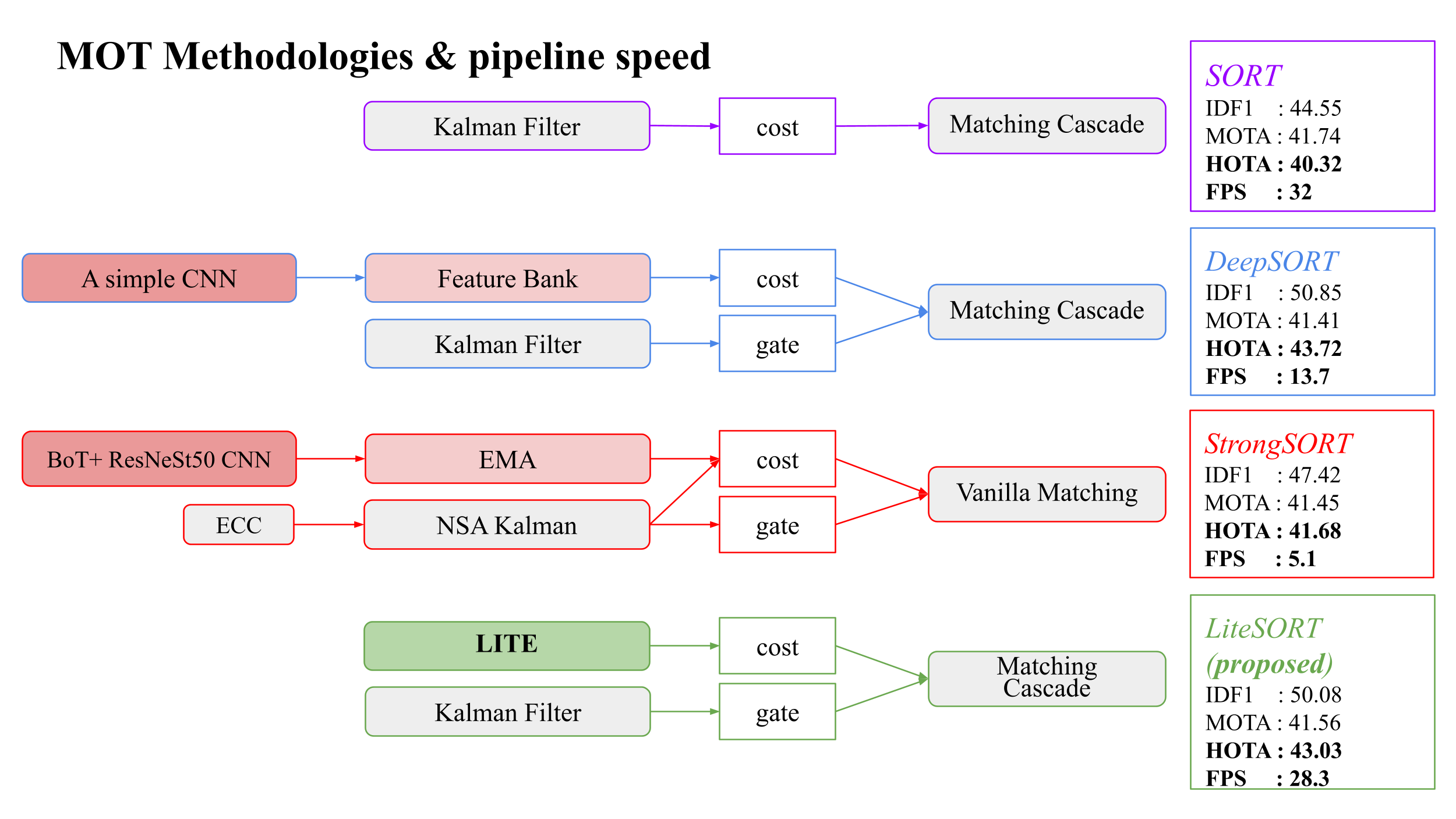}}
\caption{Overview of SORT family Trackers, their methodologies, and comparison on the MOT17 train set. The figure shows the simplicity of LiteSORT (LITE+DeepSORT) as it replaces the external ReID CNN of DeepSORT with detection feature maps via the LITE paradigm.}
\label{fig:methodology}
\end{figure*}

This paper introduces Lightweight Integrated Tracking-Feature Extraction (LITE), a novel paradigm that integrates appearance feature extraction into the tracking pipeline using a standard YOLOv8m detector. LITE:DeepSORT is the simplest implementation of LITE, speeding up the classic DeepSORT. LITE:DeepSORT achieves a Higher Order Tracking Accuracy~\cite{luiten2020IJCV} (HOTA) score of 43.03 at 28 FPS on the MOT17 benchmark Fig.\ref{fig:methodology}, doubling the speed of DeepSORT \cite{Wojke2017simple} and quintupling that of StrongSORT \cite{du2023strongsort}, while maintaining accuracy. This makes LITE suitable for real-time applications needing ReID-equipped tracking components, as in Action Detection \cite{meva_challenge,yu2022argus++}. Fig.\ref{fig:methodology} shows methodology when LITE is applied to DeepSORT.
This paper also proposes a comprehensive evaluation framework to benchmark the entire tracking pipeline, bridging the gap between reported and practical performance, especially in real-time applications. This framework highlights common pitfalls in current evaluations, ensuring a more robust assessment of tracking methods.

\section{Related Work}
\label{sec:related_work}
Trackers and their components are compared in Table \ref{tab:related_work}. Pure motion-based trackers are fast but lack ReID capability, while motion and ReID-based trackers offer ReID but are slower. FairMOT integrates both but requires specific detector training. LITE applied trackers (LITE:Trackers) combine the advantages with negligible ReID computation cost, achieving high HOTA scores with a speed advantage (details in Section \ref{sec:experiments}).

\begin{table}[!h]
\centering
\caption[Related Work]{Comparison of different trackers and their components. Columns show characteristics of the ReID component for each tracker.}
\setlength{\tabcolsep}{6pt}
\scriptsize
\begin{tabular}{ l | p{3.5em} | p{5.5em} | p{5.5em} | p{5.5em} }
\toprule
\textbf{Tracker} & \textbf{Has ReID} & \textbf{No model inference cost} & \textbf{No extra training} & \textbf{Real-time (30FPS)} \\
\midrule
\multicolumn{5}{c}{\textbf{Pure Motion}} \\
\midrule
SORT \cite{bewley2016simple} & & & & \checkmark \\
OC-SORT \cite{cao2023observation} & & & & \checkmark \\
ByteTrack \cite{zhang2022bytetrack} & & & & \checkmark \\
\midrule
\multicolumn{5}{c}{\textbf{Motion and ReID}} \\
\midrule
DeepSORT \cite{wojke2018deep} & \checkmark & & & \\
StrongSORT \cite{du2023strongsort} & \checkmark & & & \\
Bot-SORT \cite{aharon2022bot} & \checkmark & & & \\
Deep-OCSORT \cite{maggiolino2023deep} & \checkmark & & & \\
\midrule
\multicolumn{5}{c}{\textbf{Integrated Motion + ReID}} \\
\midrule
FairMOT \cite{zhang2021fairmot} & \checkmark & \checkmark & & \\
\rowcolor{babyblue!20} LITE:Trackers (proposed) & \checkmark & \checkmark & \checkmark & \checkmark \\
\bottomrule
\end{tabular}
\label{tab:related_work}
\end{table}

Significant works in MOT include ByteTrack \cite{zhang2022bytetrack} and OC-SORT \cite{cao2023observation}, focusing on motion cues, and DeepSORT \cite{Wojke2017simple} and StrongSORT \cite{du2023strongsort}, integrating deep learning-based appearance features. End-to-end frameworks like MOTR \cite{zeng2022motr} offer advancements but are limited by slow inference speeds. Trackers using LITE differentiate by integrating efficient appearance feature extraction, maintaining real-time performance without additional deep learning models.


\section{Proposed Evaluation Framework}
\label{sec:evaluation_framework}

\subsection{Motivation}
Current MOT evaluation methodologies often lack alignment with practical deployment scenarios. The proposed framework aims to bridge this gap by providing a more holistic assessment. Typically, available evaluation protocols do not closely resemble practical settings. For instance, the public evaluation protocols of the MOT17 and MOT20 benchmarks involve training detectors on training sets and evaluating them on test sets. However, in practice, pre-trained and readily available detectors, such as YOLOv8 \cite{yolov8}, are often used. Therefore, trackers following the tracking-by-detection paradigm should be evaluated using pre-trained detectors to closely mimic practical use cases.

Previous literature primarily computes FPS for matching and track management \cite{du2023strongsort,zhang2022bytetrack,cao2023observation,maggiolino2023deep}. Open-source tools measure only the tracker update speed \cite{brostrom2023boxmot,ultralytics2023multiobject}, and some state-of-the-art trackers lack speed measurements entirely \cite{maggiolino2023deep}. Additionally, frame processing time varies with the density of people, detector model complexity, input image resolution, and minimum confidence threshold. There is a lack of studies highlighting how FPS changes in response to detector settings. Such knowledge allows practitioners to better balance speed and accuracy for their computer vision problems.

\subsection{Advantages of Holistic Evaluation Framework}

The proposed framework assesses the entire tracking pipeline, including preprocessing of detection, ReID, and tracking modules; inference cost of detection and ReID modules; post-processing for detection and ReID; and the tracker’s next state predictions and updates. For example, for the first frame of KITTI’s sequence "0000", YOLOv8m took 6 ms for preprocessing, 65 ms for model inference, and 243 ms for post-processing. Similar observations are expected for ReID inference.
This holistic framework is also capable of assessing tracker performance on edge devices, where ReID-based methods may exhaust GPU memory. It provides comprehensive FPS computation by including all tracking pipeline components, bridging the gap between reported and practical performance by creating a unified pipeline.

\subsection{Real-Time Processing Requirements}
To ensure a unified pipeline, the detector must run in real-time to compute detections, real-time tracking crops for ReID should be computed by CNN networks, and the matching and track management components should also run in real-time. FPS should be computed for the entire video, accounting for varying scene complexity to gauge practical tracker speed. Performance (HOTA) and speed are influenced by detector settings, confidence thresholds, and image size. The framework evaluates the entire pipeline holistically, reflecting changes to the detector or ReID modules.

Tracking-by-detection trackers typically view detection time as separate, and open-source tools only measure tracker update speed \cite{aharon2022bot,du2023strongsort,brostrom2023boxmot}. Some state-of-the-art trackers lack speed measurements \cite{maggiolino2023deep}. Practical tracking pipeline speed is crucial for real-time applications and resource-constrained scenarios. Frame processing time varies with the density of people, detector settings, input image resolution, minimum confidence threshold, and tracker settings (initialization hits, expiration age, matching thresholds). To assess real-time speed and performance of the tracking pipeline, the proposed framework compares the speed of the entire pipeline, including preprocessing, inference, post-processing, and tracker prediction and update.

Sample FPS measurements for the SORT tracker are shown in Table \ref{tab:fps_of_sequences}, indicating variability within sequences.

\begin{table}[!htbp]
\centering 
\caption{Processing time and FPS for each video. Measurements are for the SORT tracker.}
\setlength{\tabcolsep}{10pt}
\scriptsize
\begin{tabular}{ l | l | p{10em} | l }
\toprule
\textbf{Index} & \textbf{Video} & \textbf{Total Video Sequence Processing Time (s)} & \textbf{FPS} \\
\midrule
\multicolumn{4}{c}{\textbf{MOT17 Dataset}} \\
\midrule
1 & MOT17-02-FRCNN & 17.5 & 34.5 \\
2 & MOT17-04-FRCNN & 32.1 & 32.8 \\
3 & MOT17-05-FRCNN & 19.95 & 42.1 \\
4 & MOT17-09-FRCNN & 15.38 & 34.3 \\
5 & MOT17-10-FRCNN & 18.25 & 35.9 \\
6 & MOT17-11-FRCNN & 24.77 & 36.5 \\
7 & MOT17-13-FRCNN & 21.62 & 34.8 \\
\bottomrule
\end{tabular}
\label{tab:fps_of_sequences}
\end{table}

\section{LITE}
\label{sec:LITE:DeepSORT}

Lightweight Integrated Tracking-Feature Extraction (LITE) offers an efficient method for obtaining ReID features. To demonstrate its effectiveness, LITE is applied to DeepSORT, DeepOC-SORT, and BoTSORT, showcasing a speed advantage while maintaining accuracy. Notably, LITE can be integrated into any ReID-based tracking approach, similar to the BYTE paradigm \cite{zhang2022bytetrack}.

Real-time multi-object tracking requires rapid and precise association of detected objects across video frames. Traditional ReID-equipped tracking-by-detection methods, such as DeepSORT \cite{Wojke2017simple}, BotSORT, DeepOCSORT, and StrongSORT \cite{aharon2022bot,maggiolino2023deep,du2023strongsort}, rely on separate networks for object detection and appearance features, resulting in slower performance. LITE addresses this by extracting appearance features within the object detection pipeline, significantly speeding up the process without sacrificing accuracy.

\begin{figure*}[!h]
\centering
\includegraphics[width=\textwidth]{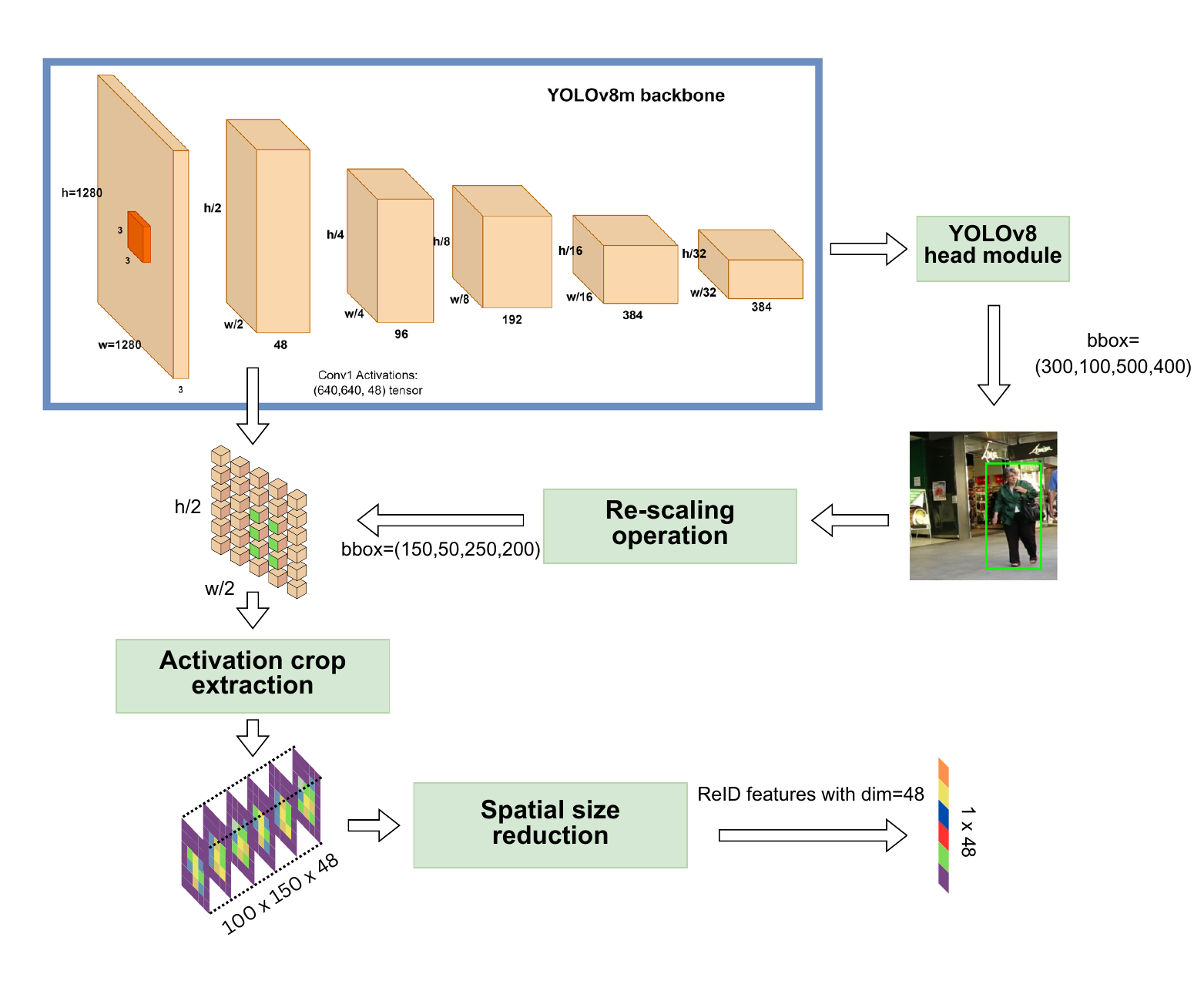}
\caption{Efficient ReID feature extraction via the LITE paradigm.}
\label{fig:novel_reid_extraction_algorithm}
\end{figure*}

LITE introduces a feature extraction mechanism that operates concurrently with object detection, harvesting appearance features from intermediate layers of the detection network as described in Fig.~\ref{fig:novel_reid_extraction_algorithm}. Using YOLOv8 \cite{yolov8}, the first convolutional layer provides a feature map with 48 channels and half the spatial resolution of the image ($\tfrac{h}{2} \times \tfrac{w}{2}$) where $h$ and $w$ correspond to input image resolution. This low-resolution, high-channel representation serves as the appearance feature map.
Following detection and non-maximum suppression (NMS), detected bounding boxes are mapped to the downscaled resolution of the appearance feature map. These regions are cropped from the feature map to produce compact appearance descriptors. Instead of utilizing the complete feature stack ($w_{\text{crop}} \times h_{\text{crop}} \times 48$), an average across channels is computed to achieve a consistent, simplified yet effective representation ($d=48$). Here \( w_{\text{crop}} \) and \( h_{\text{crop}} \) can vary and they are the scaled representation of bounding box detections. In other words, they are re-scaled to spatial resolution of corresponding conv layer activations. In the Fig.~\ref{fig:novel_reid_extraction_algorithm} \( w_{\text{crop}} \) and \( h_{\text{crop}} \) are 100 and 150 respectively.

LITE:DeepSORT is the simplest implementation of LITE, modifying the ReID computation stage of DeepSORT. It integrates appearance feature extraction directly into the tracking pipeline, eliminating the need for an external ReID model. This approach reduces computational overhead while maintaining high tracking accuracy.

While LITE applied trackers demonstrate the practicality of LITE, future research could explore higher-dimensional embeddings and integrate features from multiple levels of the detection network. That is similar to a feature pyramid network (FPN) \cite{lin2017feature}, enhancing the discriminative capability of the appearance features.

\section{Experiment Setting}
\label{sec:setting}

\subsection{Datasets}
\begin{figure}[!h]
\centering
\includegraphics[width=\textwidth]{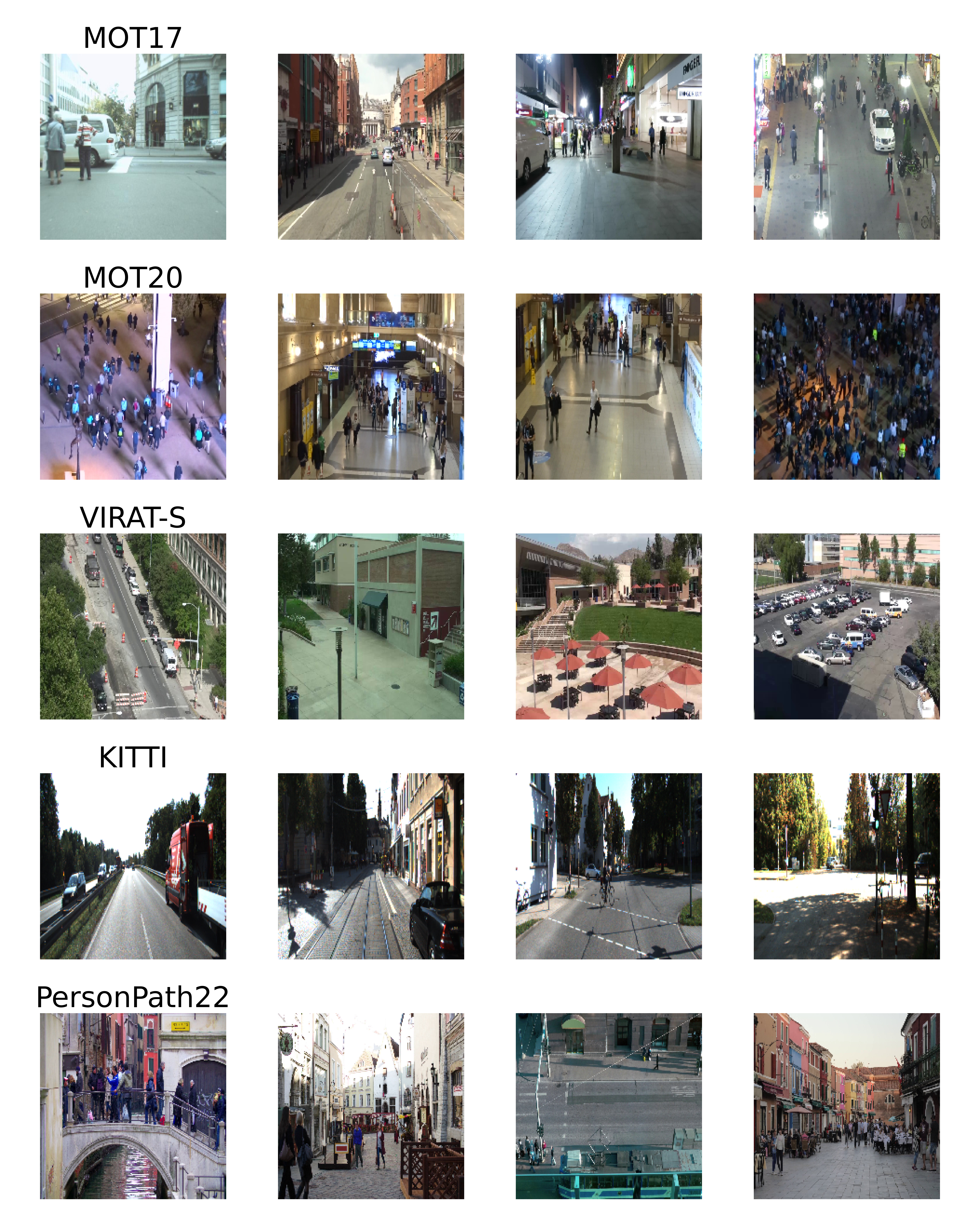}
\caption{Sample frames from various datasets used in the analysis.}
\label{fig:dataset_samples}
\end{figure}

The datasets used in the experiments include MOT17 \cite{MOT17}, which consists of 21 sequences with a resolution of 1920x1080 and an average of 650 frames (SD 200), serving as a standard multi-object tracking dataset. MOT20 \cite{MOT20} includes 8 sequences at 1920x1080 resolution, with an average of 800 frames (SD 250), presenting challenges in crowded scenes. The KITTI dataset \cite{geiger2013vision} features 50 sequences at 1242x375 resolution, averaging 120 frames (SD 30), and is used for evaluating autonomous driving scenarios. The VIRAT-S dataset \cite{alikhanov2023online} comprises 100 sequences with a resolution of 1280x720 and an average of 1000 frames (SD 300), designed for fast tracking in action detection applications. Lastly, PersonPath22 \cite{personpath22} contains over 100 sequences with various resolutions, averaging 500 frames (SD 150), and includes diverse scenarios such as indoor, outdoor, and mobile environments.
Dataset characteristics are shown in Table~\ref{tab:datasets}. MOT17 and MOT20 datasets include both train and test sequences. However, ground truth for test sequences are unavailable except for submission through portal \cite{MOT17,MOT20}. Visual samples of datasets are provided in Fig. \ref{fig:dataset_samples}.

\begin{table}
\centering
\caption{Overview of datasets used in experiments. The average and standard deviation of video length (i.e., number of frames) are shown. For MOT17 and MOT20, only training sequences are included.}
\setlength{\tabcolsep}{4pt}
\scriptsize
\begin{tabular}{ p{1.8cm} | p{3.2cm} | p{1.6cm} | p{2cm} | p{1.8cm} }
\toprule
\textbf{Dataset} & \textbf{Description} & \textbf{Resolution} & \textbf{Frames} & \textbf{\#Sequences} \\
\midrule
\textbf{MOT17} & Multi-object tracking & 1920x1080 & 650 ± 200 & 7 \\
\textbf{MOT20} & Crowded scenes & 1920x1080 & 800 ± 250 & 4 \\
\textbf{KITTI} & Autonomous driving & 1242x375 & 120 ± 30 & 21 \\
\textbf{VIRAT-S} & Action detection & 1280x720 & 1000 ± 300 & 100 \\
\textbf{PersonP22} & Diverse scenarios & Various & 500 ± 150 & 98 \\
\bottomrule
\end{tabular}
\label{tab:datasets}
\end{table}

\subsection{Implementation Details}

Experiments use two code repositories, first is StrongSORT \cite{du2023strongsort_code} selected to implement all SORT-like trackers, which are relatively simple and contain less settings and parameters. This is achieved by inhering the author's implementation of StrongSORT and DeepSORT. With simple adaptation add SORT. To add LITE:DeepSORT, replace external ReID module used in DeepSORT with LITE to obtain ReID features without external inference or pre-, post-processing steps.
While, these code is sufficient to show the strengths of LITE, more recent trackers are also added by adopting the repository BoxMOT (also known as yolo\_tracking) \cite{brostrom2023boxmot} which contains trackers such as ByteTrack, OC-SORT, DeepOC-SORT, BoTSORT. Both repository's code is adjusted to follow proposed evalation framework requirements such as holistic evaluation, real-time tracking. 

Experiments use YOLOv8m for real-time application, with a confidence threshold of 0.25 and image resolution of 1280. Despite some state-of-the-art trackers using larger architectures for better accuracy, YOLOv8m is chosen for its real-time tracking capabilities.
For MOT17, the FRCNN version of ground truth is used. For KITTI and VIRAT-S, only pedestrian and person classes are evaluated. The HOTA metric is used to evaluate tracker accuracy.
Experiments conducted on: Intel Core i9-12900K at 5.2 GHz, NVIDIA GeForce RTX 3090 with 24 GB VRAM, 64 GB DDR4 at 3200 MHz.
FPS benchmarks measured with no other significant processes running, using the first video sequence of each dataset.

\subsection{Evaluation Metrics}
Traditional metrics like Multi-Object Tracking Accuracy (MOTA) and Identification F1 Score (IDF1) have limitations that can skew tracking system evaluation. MOTA emphasizes detector accuracy, potentially misleading results when detection is challenging. IDF1 rewards association capabilities but can overlook detection accuracy. MOTA is computed based on False Positives (FP), False Negatives (FN) and Identity Switches (IDSW), while IDF1 is computed based on True Positives (IDTP), Identification False Positives (IDFP), and Identification False Negatives (IDFN).

\begin{equation}
\text{MOTA} = 1 - \frac{\text{FP} + \text{FN} + \text{IDSW}}{\text{Total Detections}}
\end{equation}

\begin{equation}
\text{IDF1} = \frac{2 \times \text{IDTP}}{2 \times \text{IDTP} + \text{IDFP} + \text{IDFN}}
\end{equation}

In contrast, HOTA balances detection accuracy and identity association quality, providing a comprehensive evaluation of a tracker's capability. By focusing on HOTA, our framework offers a holistic view of tracker performance, ensuring genuine advancements are reflected in tracking technology evaluations. AssA is the metric to measure association accuracy and DetA is for detection accuracy.

\begin{equation}
\text{HOTA} = \sqrt{\text{DetA} \times \text{AssA}}
\end{equation}


\section{Experiments}
\label{sec:experiments}

\subsection{Comparison of Trackers with Off-the-Shelf Detectors}
The experiments aim to highlight common pitfalls in prior evaluation settings and compare LITE version of ReID based trackers against their original counterparts in terms of speed.
LITE applied trackers are compared with other trackers, presenting results for HOTA, IDF1, MOTA, and FPS. Results for commonly used benchmark are shown in Table \ref{tab:tracker_comparison}, corresponding to Fig. \ref{fig:methodology}. 

\begin{figure}[!h]
\centering
\caption*{Comparative Analysis of Tracking Methods on MOT17-04-FRCNN}

\begin{subfigure}[b]{0.3\textwidth}
    \centering
    \includegraphics[width=\textwidth]{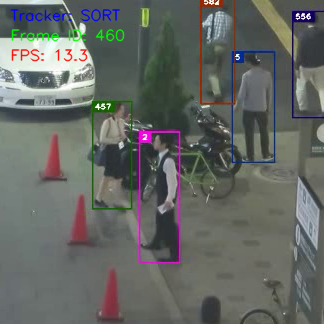}
    \caption{Frame 460}
\end{subfigure}%
\begin{subfigure}[b]{0.3\textwidth}
    \centering
    \includegraphics[width=\textwidth]{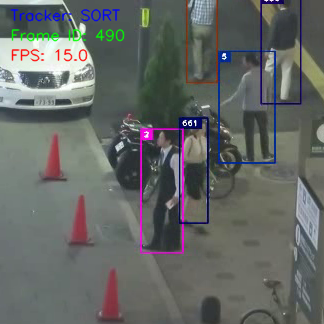}
    \caption{Frame 490}
\end{subfigure}%
\begin{subfigure}[b]{0.3\textwidth}
    \centering
    \includegraphics[width=\textwidth]{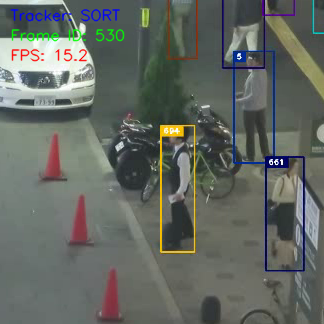}
    \caption{Frame 530}
\end{subfigure}
\caption*{SORT: ID switches after path crossing, showing the need for appearance features}
\label{fig:sort_MOT17-04-FRCNN}

\begin{subfigure}[b]{0.3\textwidth}
    \centering
    \includegraphics[width=\textwidth]{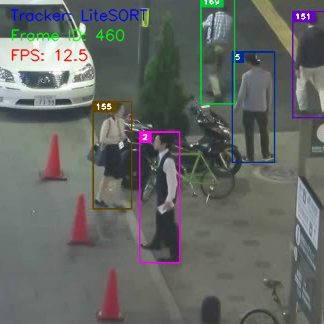}
    \caption{Frame 460}
\end{subfigure}%
\begin{subfigure}[b]{0.3\textwidth}
    \centering
    \includegraphics[width=\textwidth]{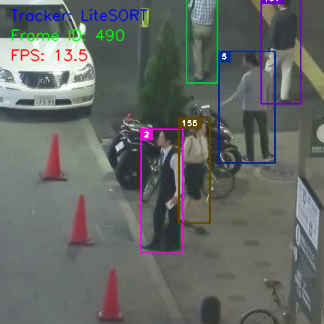}
    \caption{Frame 490}
\end{subfigure}%
\begin{subfigure}[b]{0.3\textwidth}
    \centering
    \includegraphics[width=\textwidth]{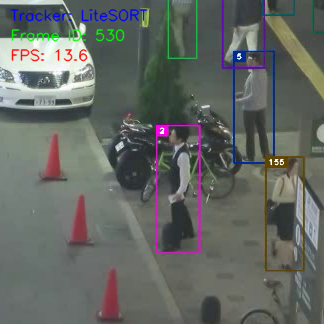}
    \caption{Frame 530}
\end{subfigure}
\caption*{LITE:DeepSORT: No ID switches after path crossing}
\label{fig:LiteSORT_MOT17-04-FRCNN}

\caption{Tracking process for a person with ID=2 (pink bounding box)}
\label{fig:MOT17-04-FRCNN}
\end{figure}

Qualitative results highlight the differences among trackers and emphasize the importance of ReID capability. Fig. \ref{fig:MOT17-04-FRCNN} illustrates the performance of each tracker. For more qualitative comparisons, refer to the Supplementary Materials.

As mentioned in proposed evaluation framework and will be shown in later experiments, ranking difference between HOTA is not sufficient from these experiments alone. Hence, avoid making any conclusion here. However, visible comparison is classic trackers such as DeepSORT are still competitive when proposed fair evaluation framework is applied.

\begin{table}[!h]
\centering
\caption{Performance comparison of different trackers on MOT17.}
\setlength{\tabcolsep}{8pt} 
\scriptsize
\begin{tabular}{ l | p{2.5em} p{2.5em} p{2.5em} p{2.5em} p{2.5em} p{2.5em} }
\toprule
 & \multicolumn{6}{c}{\textbf{MOT17}} \\ 
\cline{2-7}
\addlinespace[0.4em]
\textbf{Tracker} & \textbf{HOTA}$\uparrow$ & \textbf{IDF1}$\uparrow$ & \textbf{MOTA}$\uparrow$ & \textbf{AssA}$\uparrow$ & \textbf{DetA}$\uparrow$ & \textbf{FPS}$\uparrow$ \\ 
\midrule
StrongSORT  & 41.7 & 47.4 & 41.5 & 40.4 & 43.6 & 5.1    \\ 
BoTSORT     & 40.9 & 46.0 & 41.1 & 42.1 & 40.0 & 19.1   \\
ByteTrack   & 43.8 & 51.5 & 42.9 & 45.5 & 42.6 & 29.7   \\
OCSORT      & 43.9 & 51.0 & 41.9 & 45.2 & 43.3 & 28.8   \\
DeepOC-SORT & 43.7 & 50.7 & 43.0 & 45.8 & 42.2 & 10.6   \\
\rowcolor{babyblue!20} LITE:DeepOC-SORT    & 43.7 & 50.7 & 42.9 & 45.7 & 42.2 & 27.9    \\ 
\rowcolor{babyblue!20} LITE:BoTSORT    & 40.8 & 45.9 & 41.1 & 42.0 & 40.0 & 30.7    \\ 
\midrule
SORT        & 40.3 & 44.5 & 41.7 & 39.0 & 42.4 & 32.0   \\ 
DeepSORT    & 43.7 & 50.9 & 41.4 & 43.8 & 43.0 & 13.7   \\ 
\rowcolor{babyblue!20} LITE:DeepSORT    & 43.0 & 50.1 & 41.6 & 43.4 & 43.2 & 28.3    \\ 
\bottomrule
\end{tabular}
\label{tab:tracker_comparison}
\end{table}

\subsection{Evaluation Under Various Scenarios}
To assess the robustness of the trackers, we conducted experiments across diverse scenarios using corresponding datasets. Specifically, all trackers, including ones that are using LITE, were evaluated for their speed advantage and generalizability across the MOT17, MOT20, PersonPath22, KITTI, and VIRAT-S datasets, employing the HOTA and FPS metrics.

As results shown in Table \ref{tab:performance_comparison_across_different_datasets}, LITE:DeepSORT, LITE:DeepOC-SORT, and LITE:BoTSORT maintain competitive tracking accuracy while offering a significant speed advantage. This makes applying LITE highly suitable for real-time multi-object tracking in various challenging environments. Notably, LITE applied trackers' speed advantage is particularly evident in crowded scenarios such as MOT20. It is important to mention that other ReID-based trackers employ more complex ReID modules, which results in a much higher speed increase when LITE is applied.

\begin{table}
\centering
\caption{Performance comparison of trackers across different datasets.}
\setlength{\tabcolsep}{5pt} 
\scriptsize
\resizebox{\textwidth}{!}{
\begin{tabular}{ l | p{2.5em}p{2.5em} | p{2.5em}p{2.5em} | p{2.5em}p{2.5em} | p{2.5em}p{2.5em} | p{2.5em}p{2.5em} }
\toprule
 & \multicolumn{2}{c|}{\textbf{MOT17}} & \multicolumn{2}{c|}{\textbf{MOT20}} & \multicolumn{2}{c|}{\textbf{PersonPath22}} & \multicolumn{2}{c|}{\textbf{KITTI}} & \multicolumn{2}{c}{\textbf{VIRAT-S}} \\
\cline{2-11}
\addlinespace[0.4em]
\textbf{Tracker} & \textbf{HOTA}$\uparrow$ & \textbf{FPS}$\uparrow$ & \textbf{HOTA}$\uparrow$ & \textbf{FPS}$\uparrow$ & \textbf{HOTA}$\uparrow$ & \textbf{FPS}$\uparrow$ & \textbf{HOTA}$\uparrow$ & \textbf{FPS}$\uparrow$ & \textbf{HOTA}$\uparrow$ & \textbf{FPS}$\uparrow$ \\
\midrule
StrongSORT  & 41.7 & 5.1 & 24.8 & 2.7 & 38.0 & 5.9 & 44.0 & 23.6 & 33.4 & 21.8    \\
BoTSORT     & 40.9 & 19.1 & 20.8 & 14.3 & 39.1 & 18.4 & 33.7 & 22.5 & 31.1 & 30.0 \\
ByteTrack   & 43.8 & 29.7 & 25.2 & 24.4 & 40.5 & 27.0 & 44.9 & 27.2 & 32.7 & 37.2 \\
OCSORT      & 43.9 & 28.8 & 25.2 & 24.2 & 40.3 & 26.6 & 43.9 & 25.5 & 31.9 & 37.0 \\
DeepOC-SORT & 43.7 & 10.6 & 24.9 & 7.2 & 39.9 & 13.9 & 43.7 & 18.7 & 31.7 & 24.0  \\
\rowcolor{babyblue!20} LITE:DeepOC-SORT & 43.7 & 27.9 & 25.3 & 22.7 & 39.8 & 25.0 & 44.1 & 23.0 & 31.5 & 36.5  \\
\rowcolor{babyblue!20} LITE:BoTSORT & 40.8 & 30.7 & 21.1 & 24.2 & 39.2 & 26.4 & 33.0 & 24.1 & 31.2 & 38.0 \\
\midrule
SORT        & 40.3 & 32.0 & 20.1 & 27.0 & 35.2 & 30.8 & 41.1 & 43.1 & 28.3 & 42.7 \\
DeepSORT    & 43.7 & 13.7 & 24.8 & 5.5 & 38.3 & 15.1 & 42.6 & 38.3 & 33.6 & 33.9  \\
\rowcolor{babyblue!20} LITE:DeepSORT    & 43.0 & 28.3 & 25.2 & 23.4 & 38.0 & 26.1 & 42.8 & 40.8 & 33.7 & 40.2 \\
\bottomrule
\end{tabular}%
}
\label{tab:performance_comparison_across_different_datasets}
\end{table}

\subsection{Effect of Detector Settings on Tracking Pipelines}
\begin{figure*}[!htbp]
\centering
\begin{subfigure}{.48\linewidth}
\includegraphics[width=\linewidth]{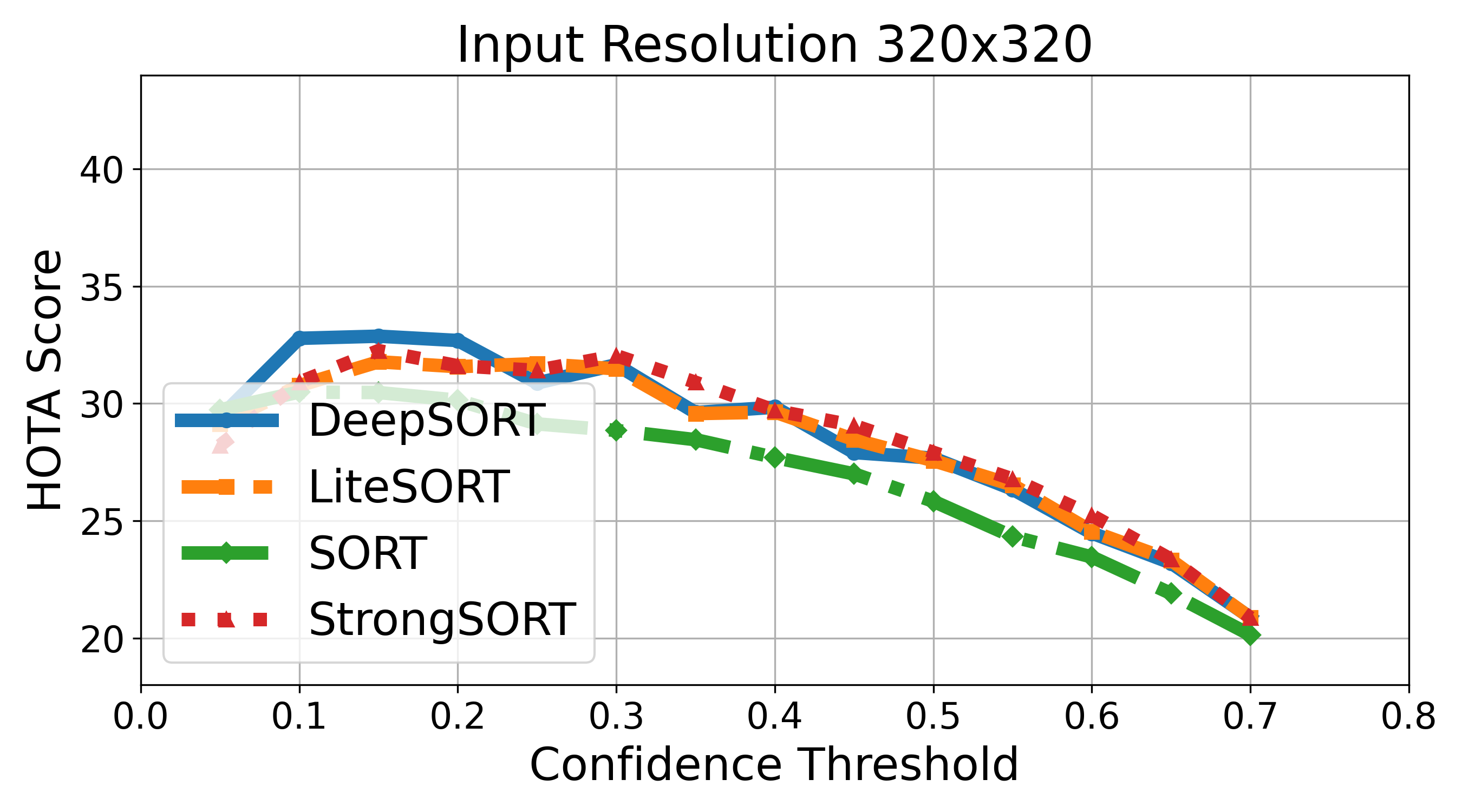}
\caption{Effect of detection settings on HOTA for resolution 320x320.}
\label{fig:hota_320x320}
\end{subfigure}%
\hfill
\begin{subfigure}{.48\linewidth}
\includegraphics[width=\linewidth]{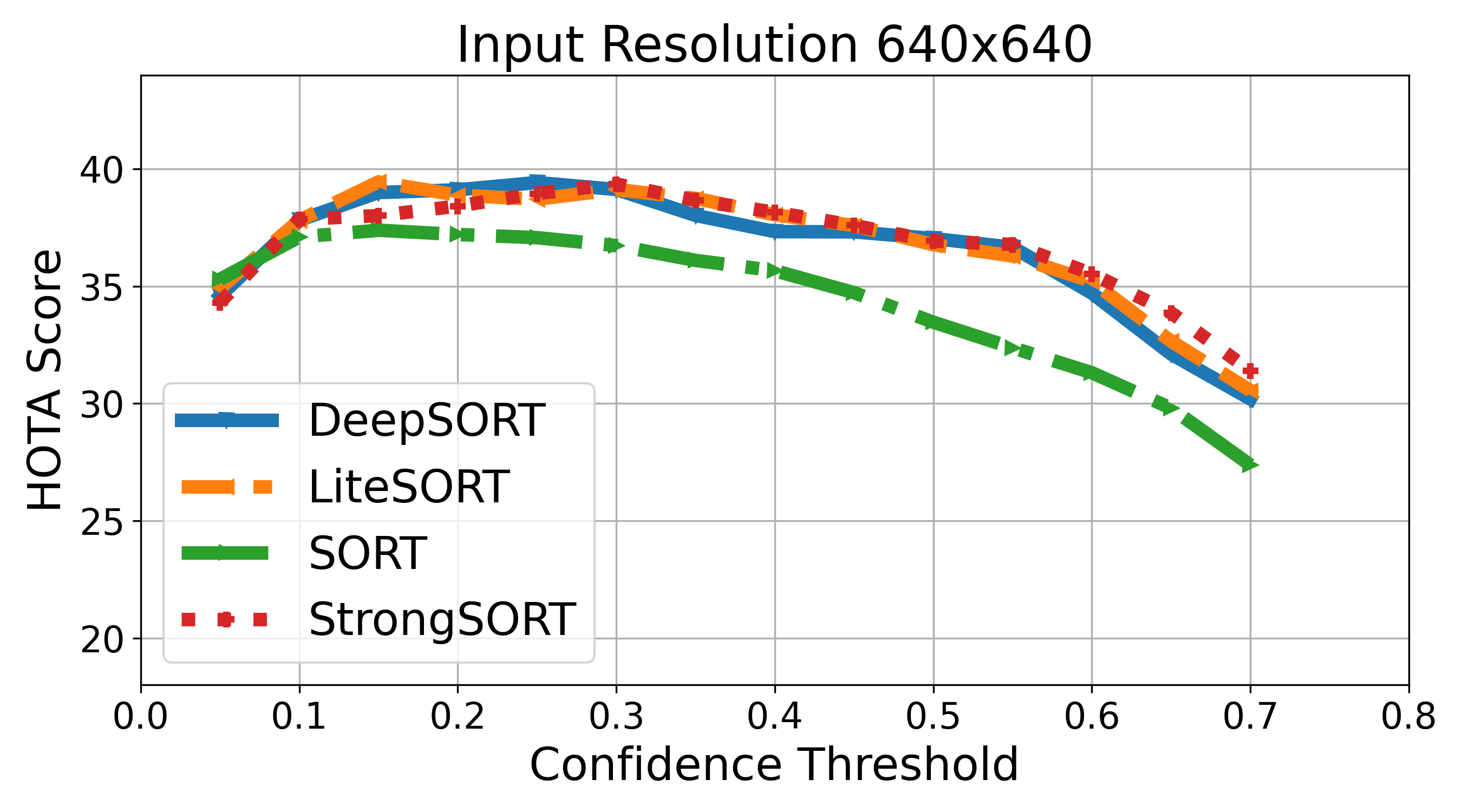}
\caption{Effect of detection settings on HOTA for resolution 640x640.}
\label{fig:hota_640x640}
\end{subfigure}
\begin{subfigure}{.48\linewidth}
\includegraphics[width=\linewidth]{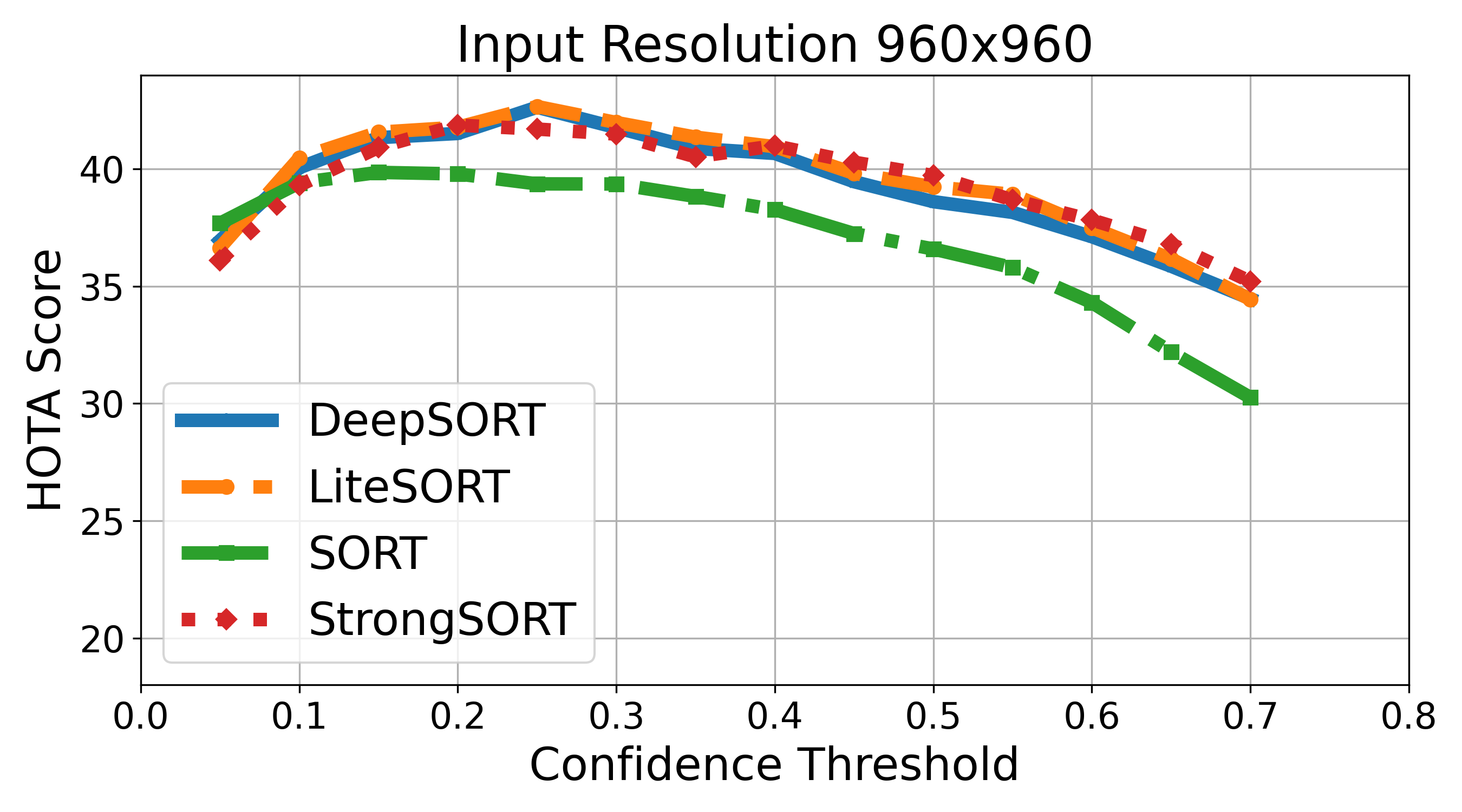}
\caption{Effect of detection settings on HOTA for resolution 960x960.}
\label{fig:hota_960x960}
\end{subfigure}%
\hfill
\begin{subfigure}{.48\linewidth}
\includegraphics[width=\linewidth]{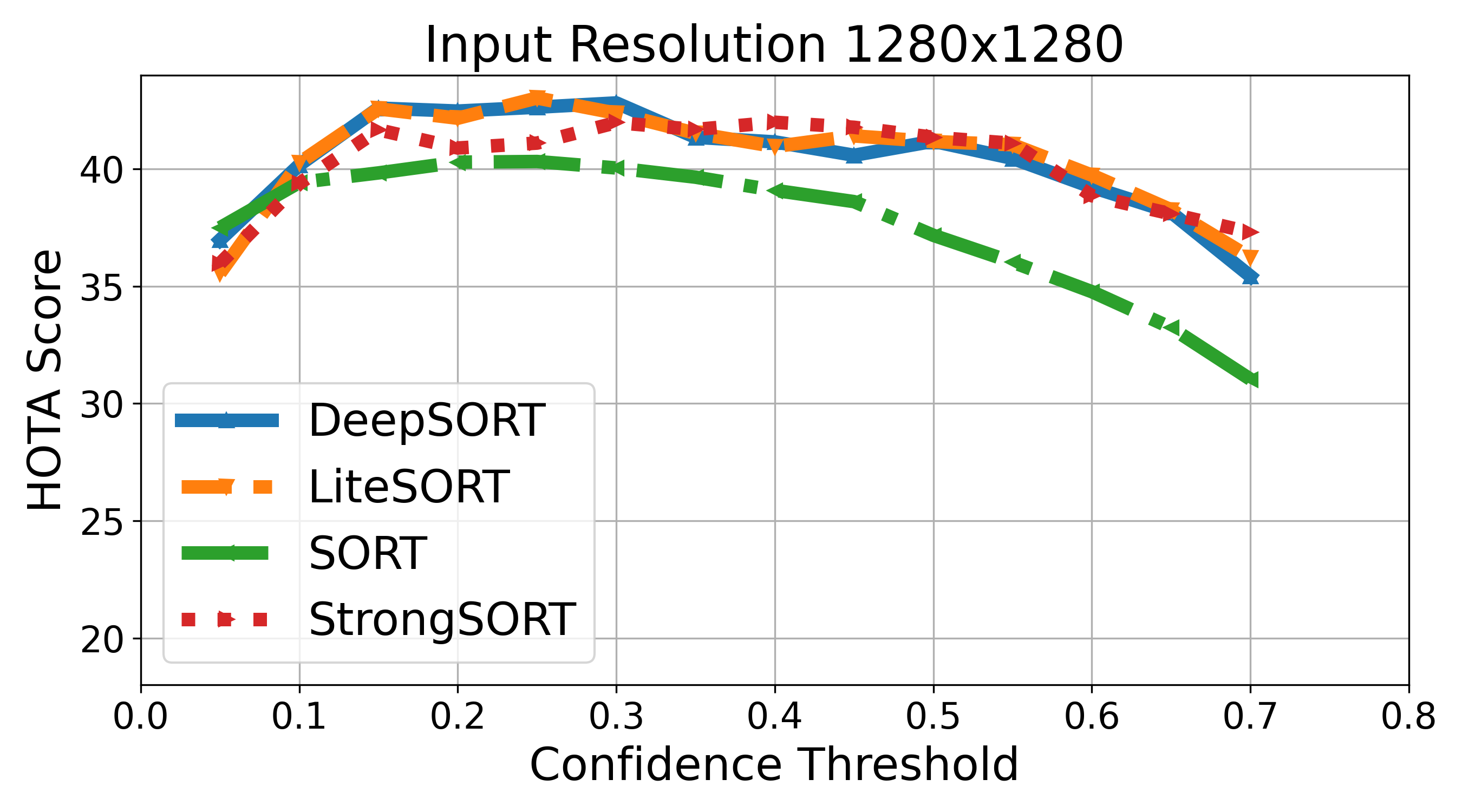}
\caption{Effect of detection settings on HOTA for resolution 1280x1280.}
\label{fig:hota_1280x1280}
\end{subfigure}
\begin{subfigure}{.48\linewidth}
\includegraphics[width=\linewidth]{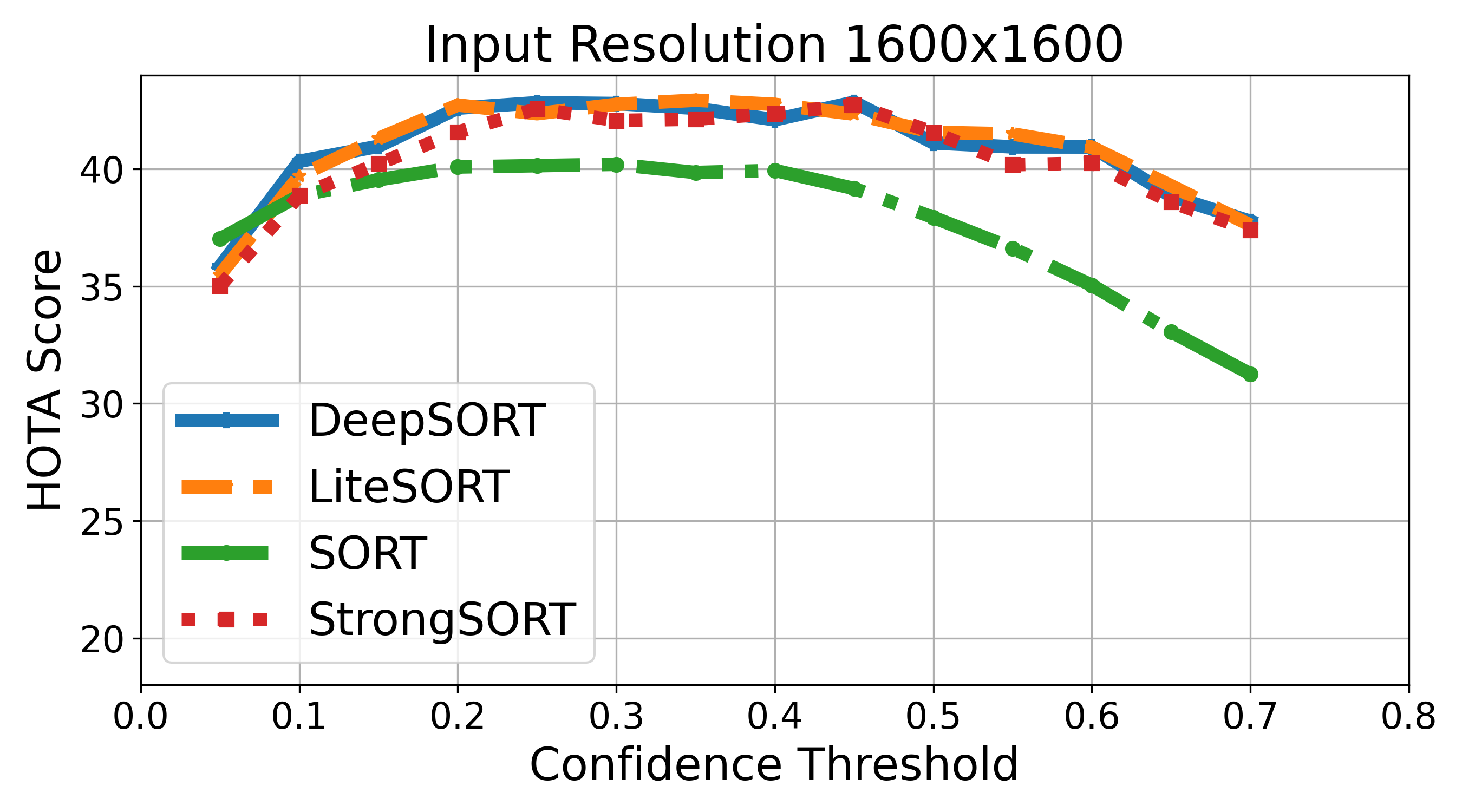}
\caption{Effect of detection settings on HOTA for resolution 1600x1600.}
\label{fig:hota_1600x1600}
\end{subfigure}%
\hfill
\begin{subfigure}{.48\linewidth}
\includegraphics[width=\linewidth]{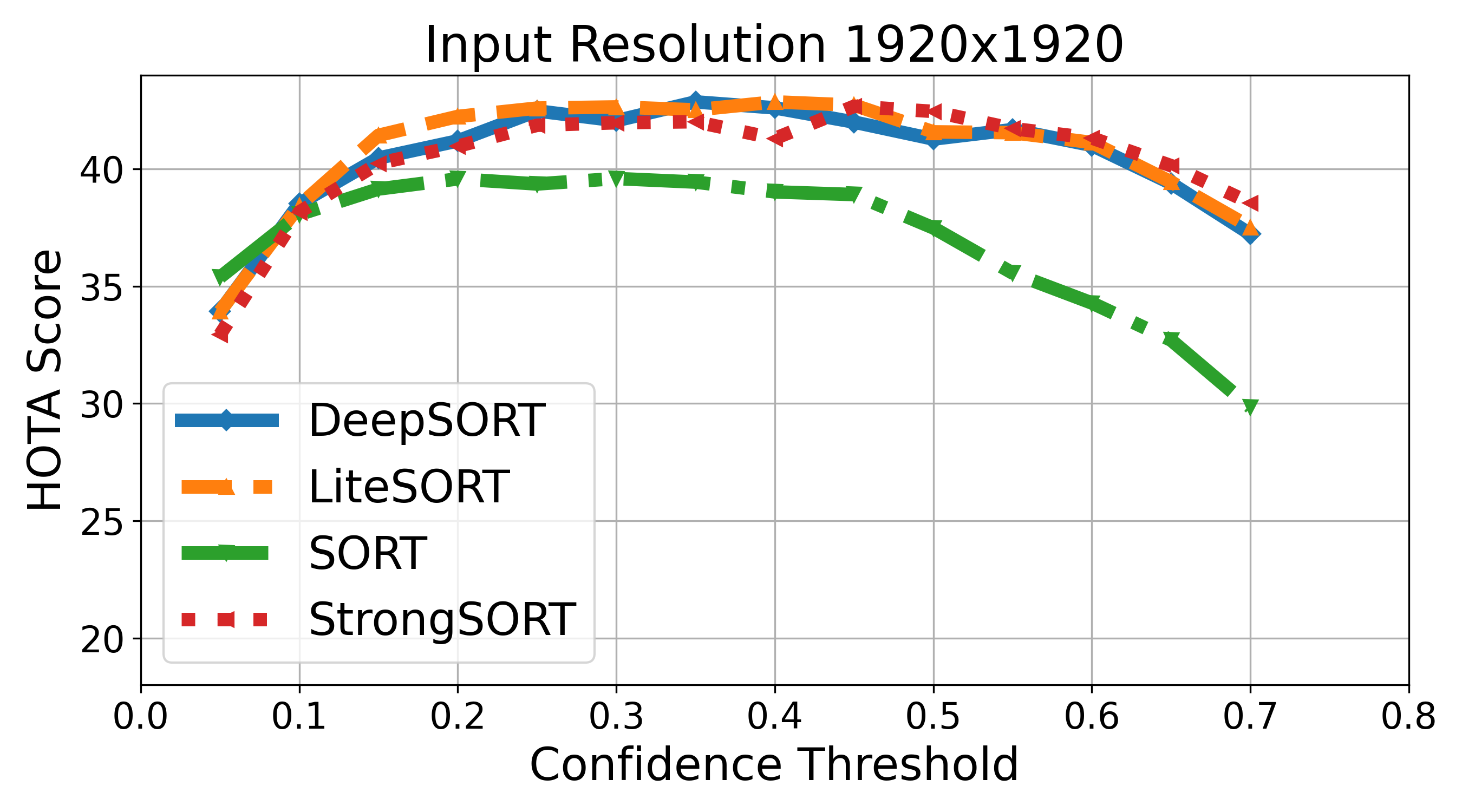}
\caption{Effect of detection settings on HOTA for resolution 1920x1920.}
\label{fig:hota_1920x1920}
\end{subfigure}
\begin{subfigure}{.48\linewidth}
\includegraphics[width=\linewidth]{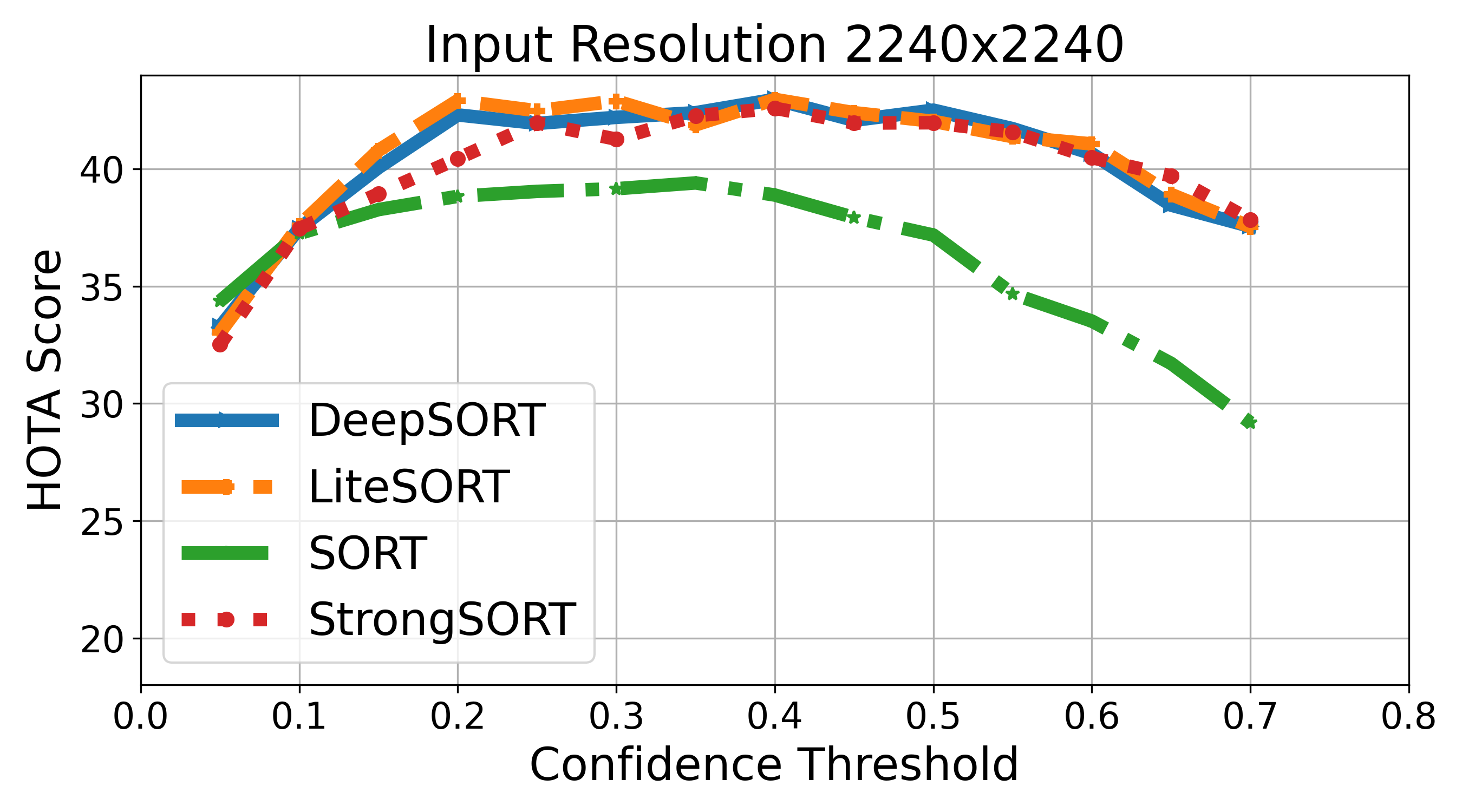}
\caption{Effect of detection settings on HOTA for resolution 2240x2240.}
\label{fig:hota_2240x2240}
\end{subfigure}%
\hfill
\begin{subfigure}{.48\linewidth}
\includegraphics[width=\linewidth]{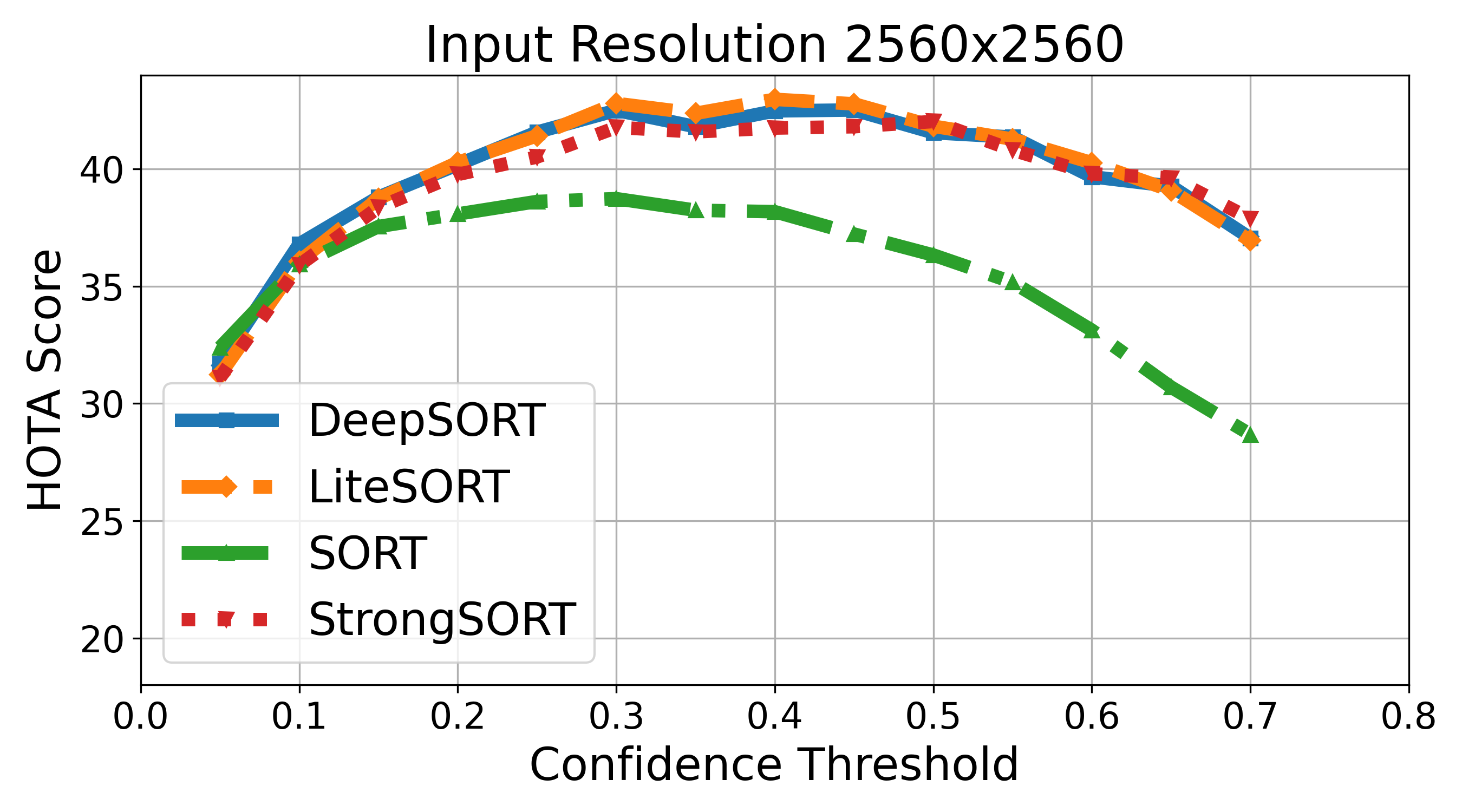}
\caption{Effect of detection settings on HOTA for resolution 2560x2560.}
\label{fig:hota_2560x2560}
\end{subfigure}
\caption{Comparative analysis of detection settings' impact on HOTA scores across different input resolutions. Experiments conducted on the MOT17 dataset. LiteSORT is an alias for LITE:DeepSORT.}
\label{fig:det-settings-HOTA}
\end{figure*}

\begin{figure*}[!htbp]
\centering
\begin{subfigure}{.48\linewidth}
\includegraphics[width=\linewidth]{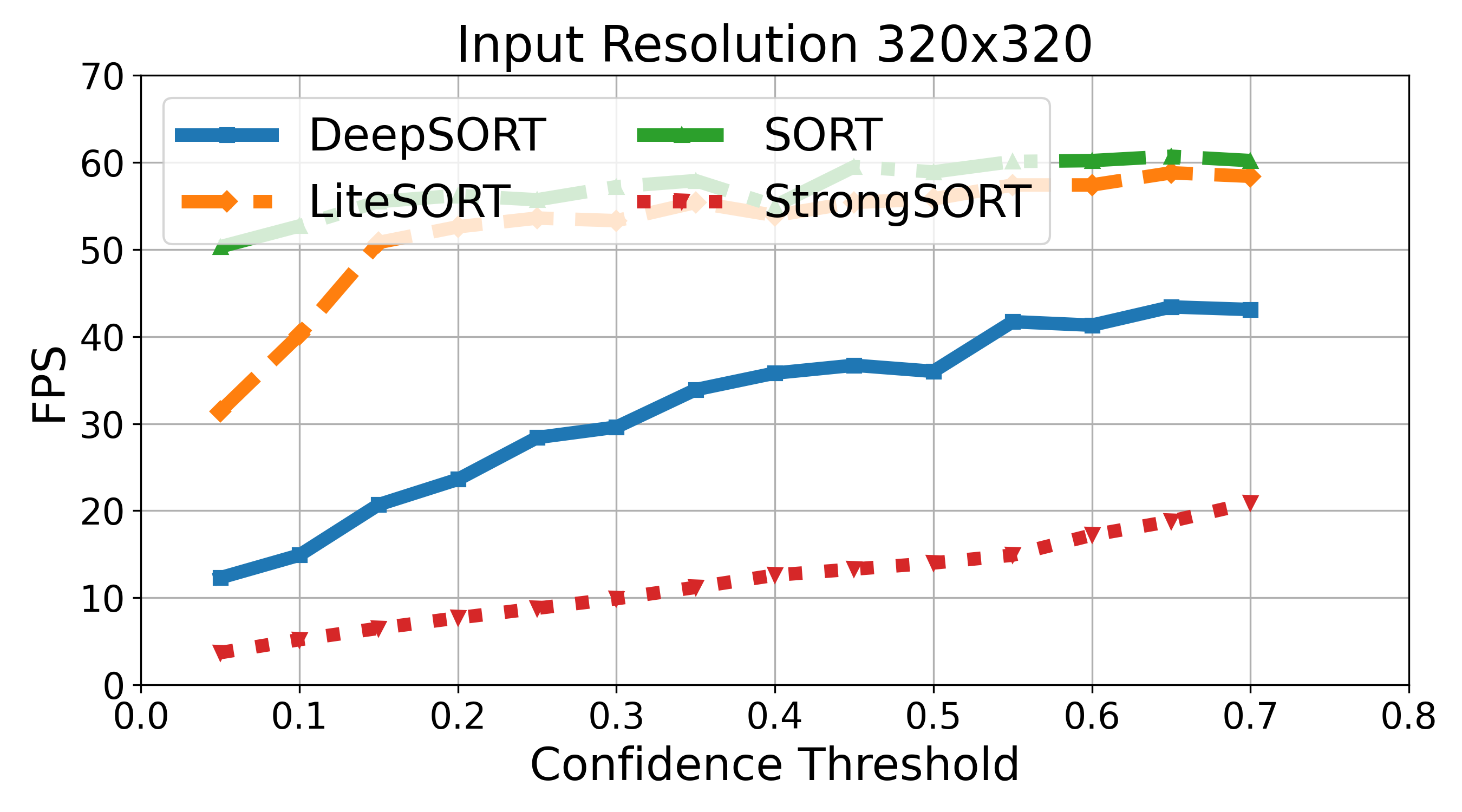}
\caption{Effect of detection settings on FPS for resolution 320x320.}
\label{fig:320x320}
\end{subfigure}%
\hfill
\begin{subfigure}{.48\linewidth}
\includegraphics[width=\linewidth]{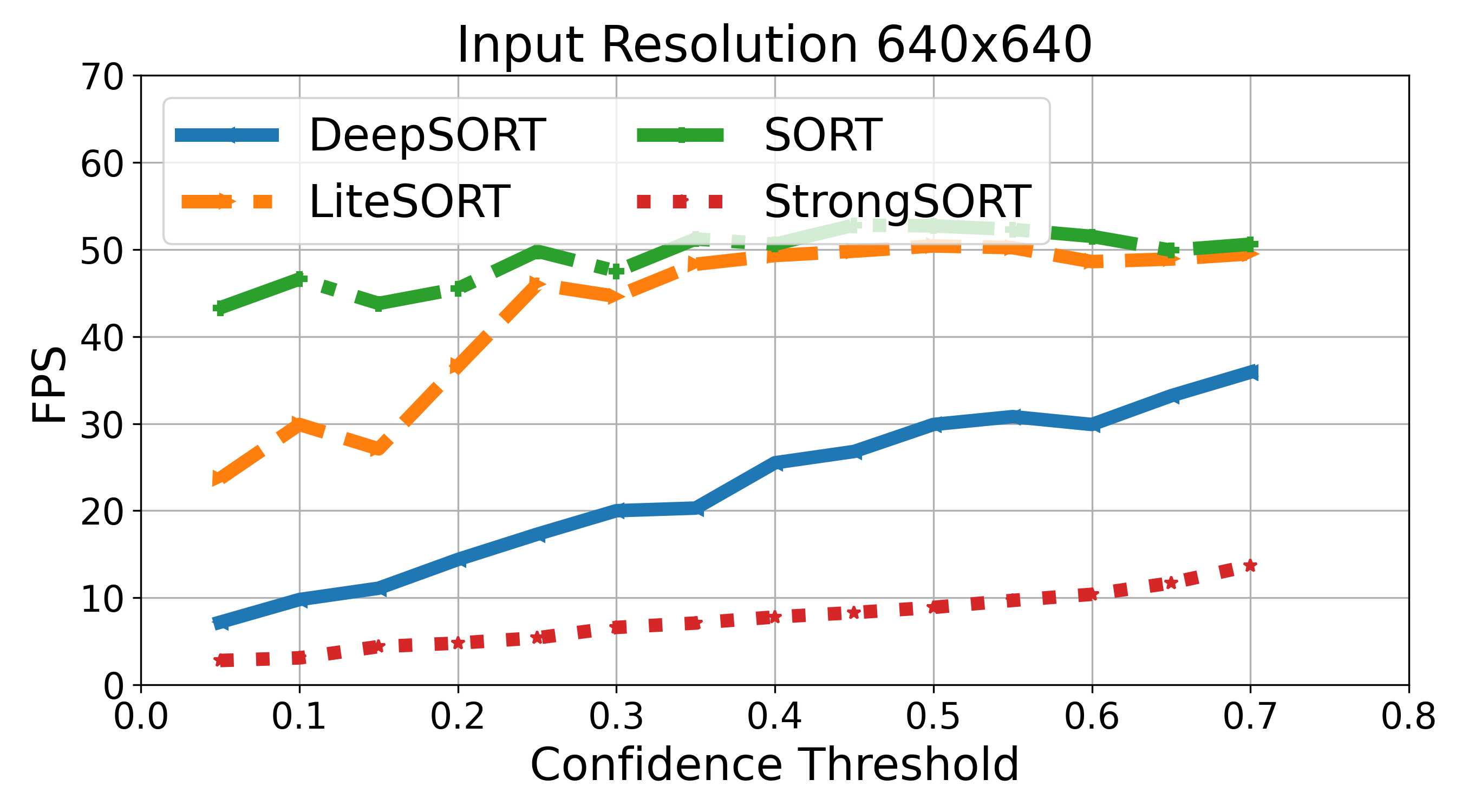}
\caption{Effect of detection settings on FPS for resolution 640x640.}
\label{fig:640x640}
\end{subfigure}
\begin{subfigure}{.48\linewidth}
\includegraphics[width=\linewidth]{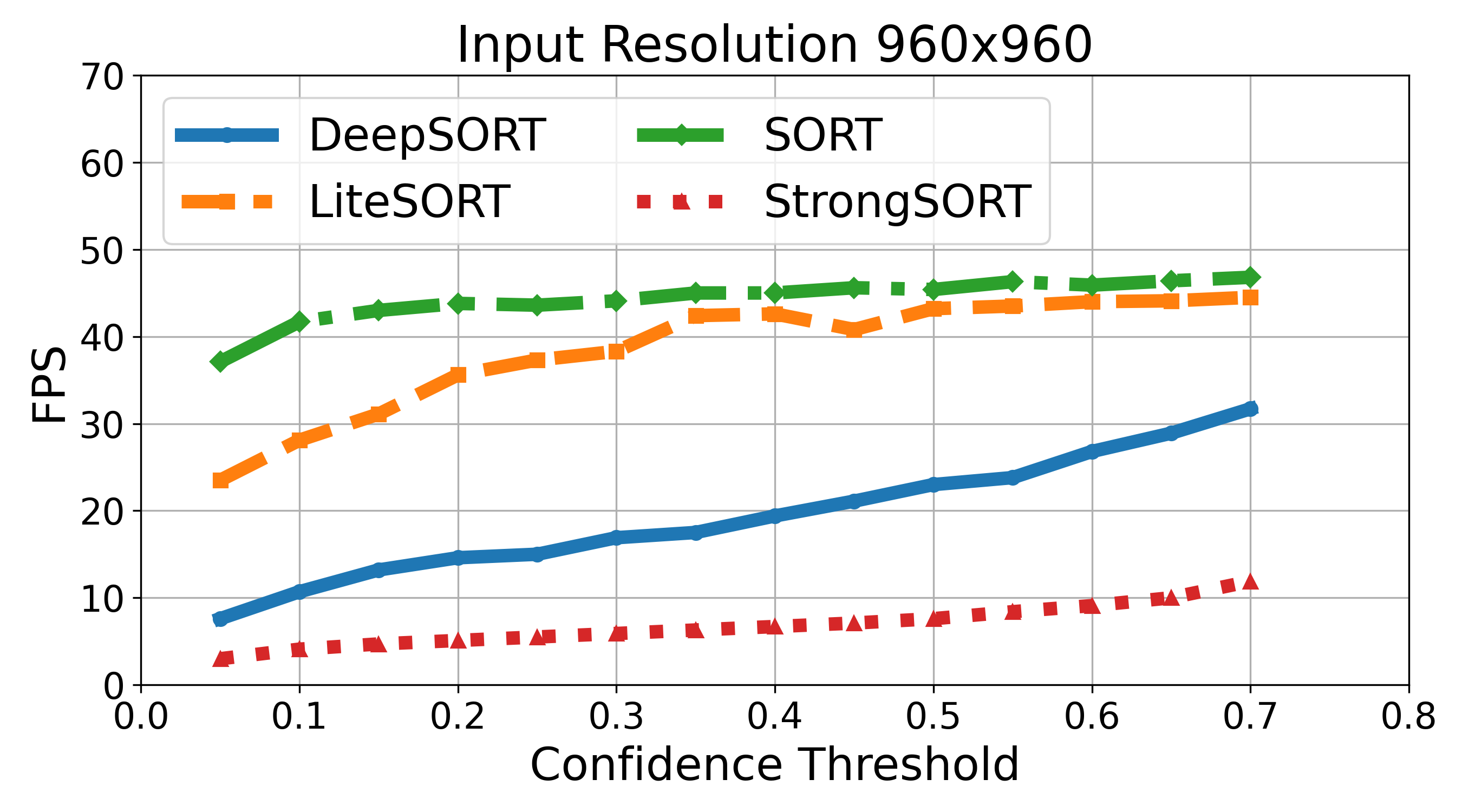}
\caption{Effect of detection settings on FPS for resolution 960x960.}
\label{fig:960x960}
\end{subfigure}%
\hfill
\begin{subfigure}{.48\linewidth}
\includegraphics[width=\linewidth]{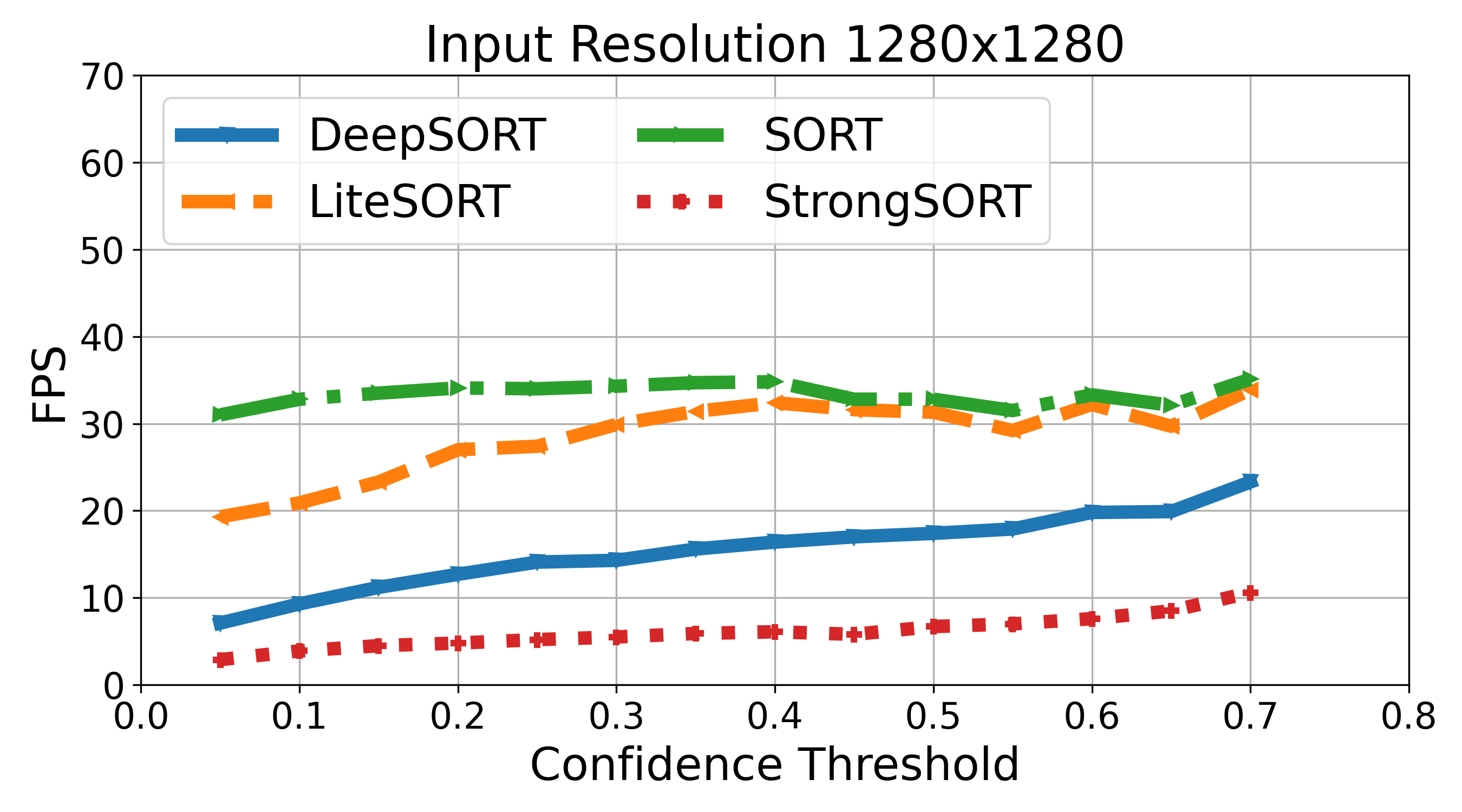}
\caption{Effect of detection settings on FPS for resolution 1280x1280.}
\label{fig:1280x1280}
\end{subfigure}
\begin{subfigure}{.48\linewidth}
\includegraphics[width=\linewidth]{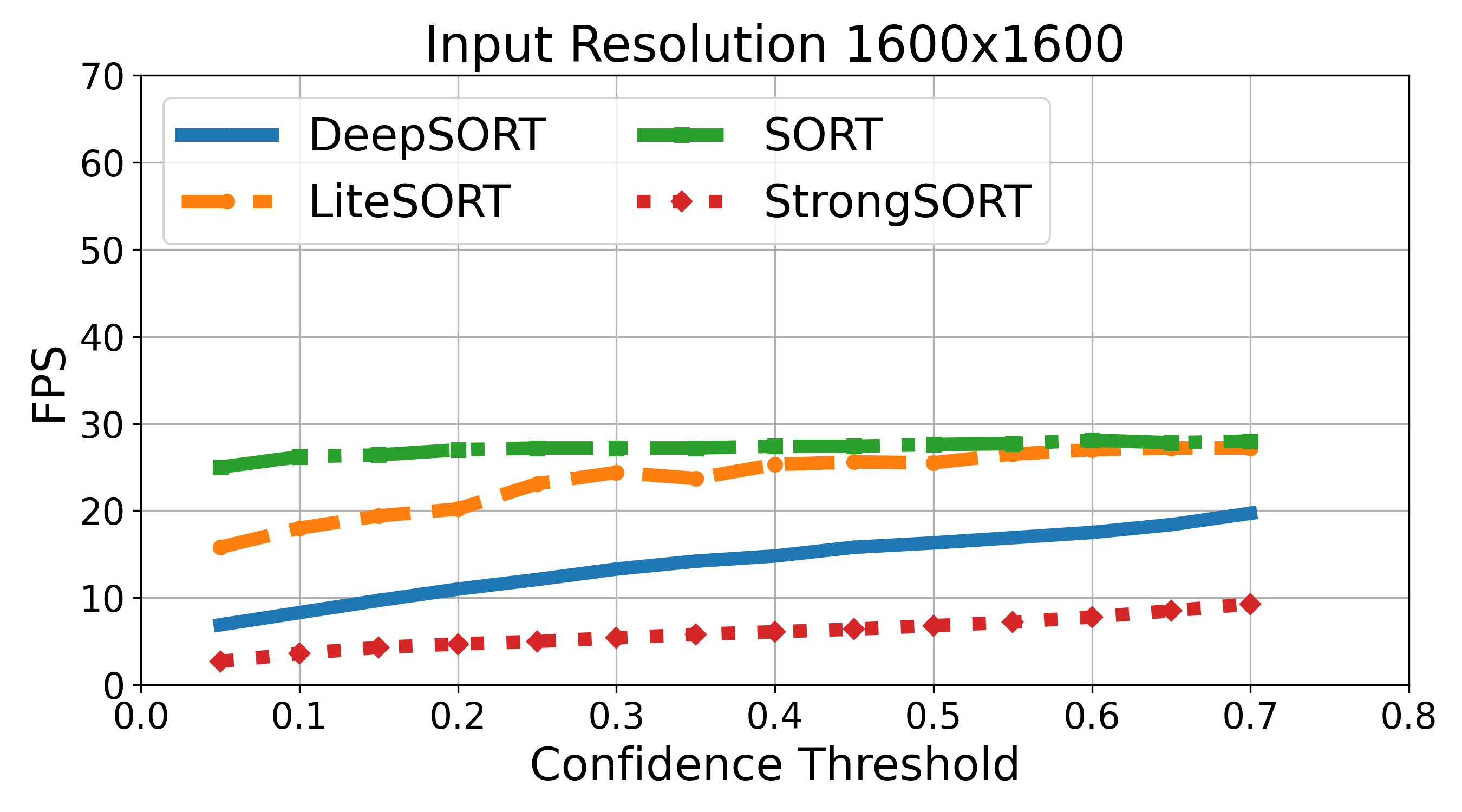}
\caption{Effect of detection settings on FPS for resolution 1600x1600.}
\label{fig:1600x1600}
\end{subfigure}%
\hfill
\begin{subfigure}{.48\linewidth}
\includegraphics[width=\linewidth]{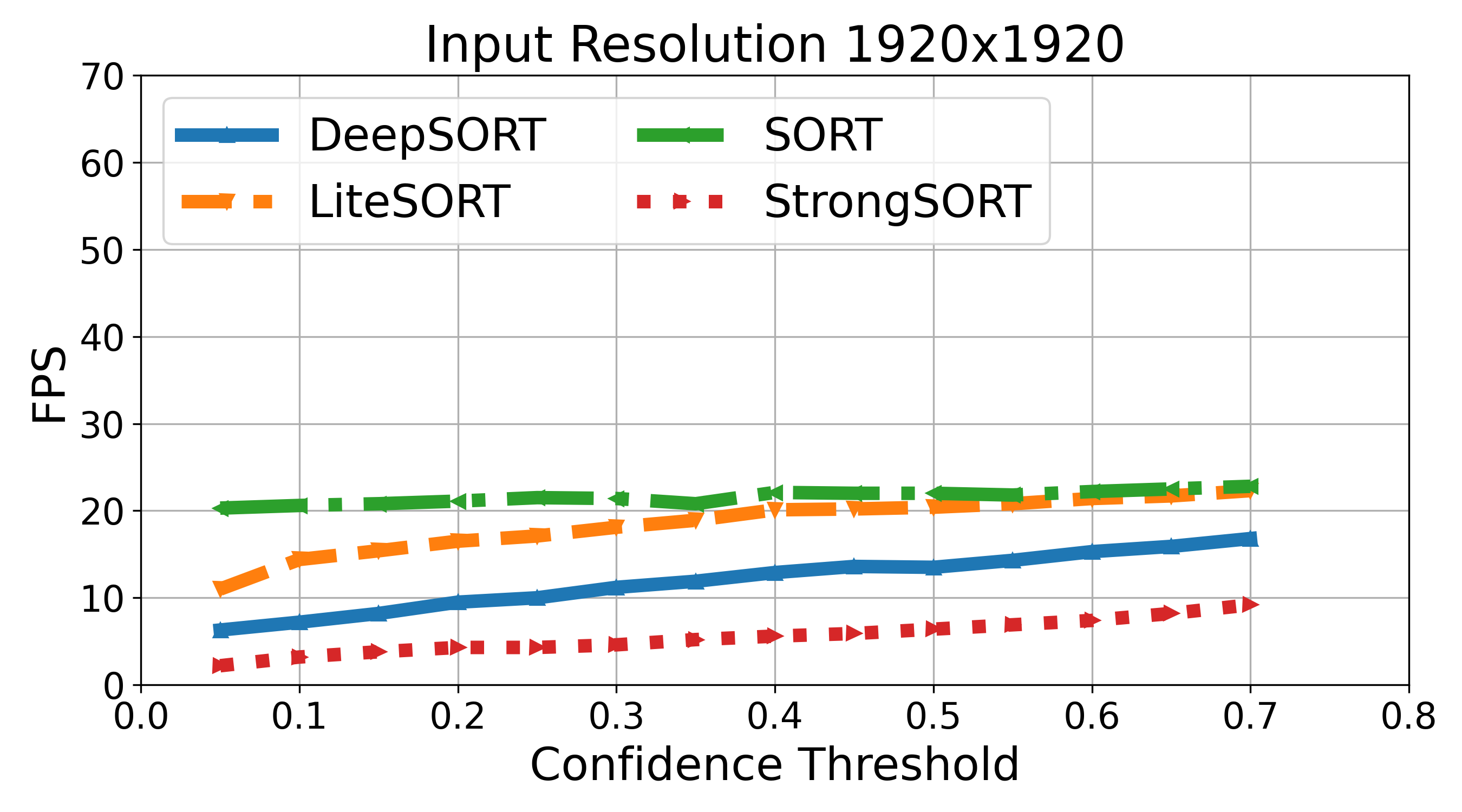}
\caption{Effect of detection settings on FPS for resolution 1920x1920.}
\label{fig:1920x1920}
\end{subfigure}
\begin{subfigure}{.48\linewidth}
\includegraphics[width=\linewidth]{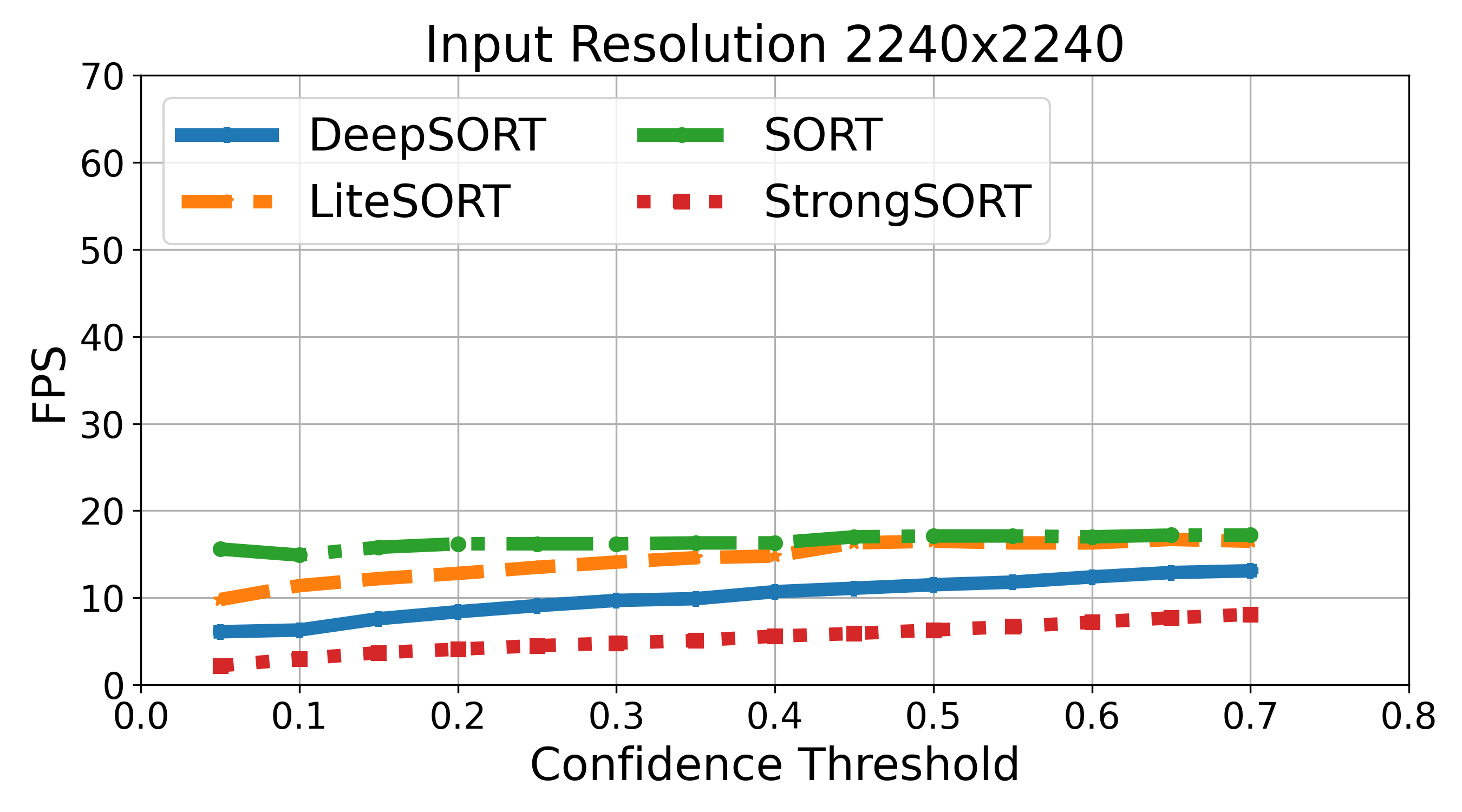}
\caption{Effect of detection settings on FPS for resolution 2240x2240.}
\label{fig:2240x2240}
\end{subfigure}%
\hfill
\begin{subfigure}{.48\linewidth}
\includegraphics[width=\linewidth]{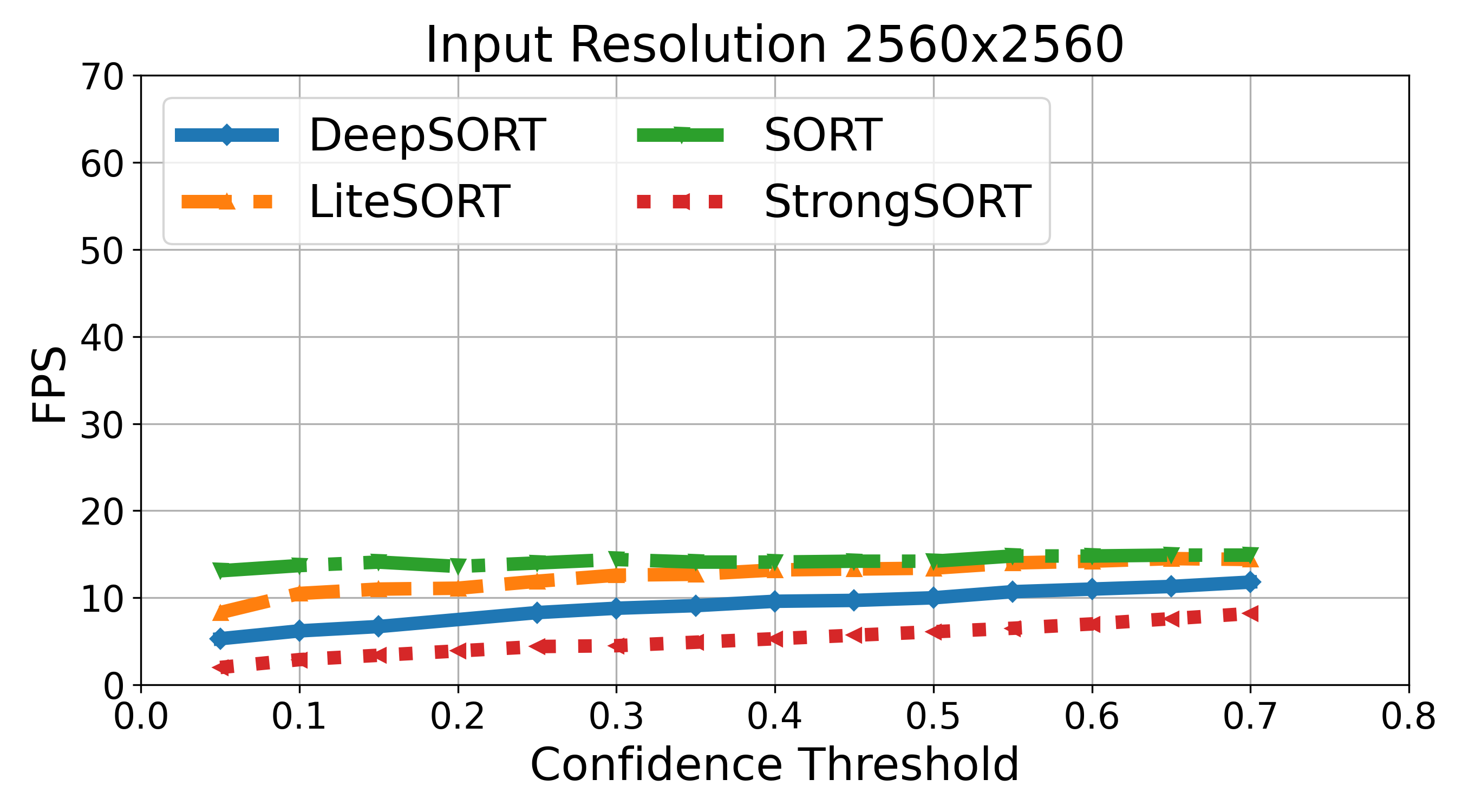}
\caption{Effect of detection settings on FPS for resolution 2560x2560.}
\label{fig:2560x2560}
\end{subfigure}
\caption{Comparative analysis of detection settings' impact on FPS scores across different input resolutions. Experiments conducted on the MOT17 dataset. LiteSORT is an alias for LITE:DeepSORT.}
\label{fig:det-settings-FPS}
\end{figure*}

The impact of different detector settings on tracking performance is analyzed. Trackers such as SORT and LITE:DeepSORT benefit from stronger detection pipelines with larger image resolutions, while DeepSORT and StrongSORT suffer from lower FPS due to slower pipelines. Figures \ref{fig:det-settings-HOTA} and \ref{fig:det-settings-FPS} illustrate the effect of detection settings on HOTA and FPS, respectively.

The main insight from these figures is that small HOTA differences (1\%-2\%) between trackers are insufficient to conclude one is superior to another. Fluctuations in ranking can occur due to various factors and are influenced by the small scale of benchmark datasets. Detector settings significantly impact performance, often more than the differences between trackers. Larger resolutions typically require higher confidence thresholds, which affect pipeline speed. Lower thresholds generate more detections, leading to more ID switches and increased pre- and post-processing costs, especially if the tracker includes a ReID component.

\section{Conclusion}
\label{sec:conclusion}

This paper presents a practical evaluation framework for assessing real-world tracker performance, including detection, ReID, tracking, and their components. Extensive evaluations highlight the shortcomings of previous methods.

LITE is introduced to obtain ReID features without training, separate inference, or pre- and post-processing, allowing ReID-based tracking with speeds close to motion-based trackers like SORT. LITE:DeepSORT, LITE:DeepOC-SORT, and LITE:BoTSORT demonstrate this effectiveness across multiple benchmarks, proving to be 2-10 times faster than their original counterparts while maintaining similar HOTA scores.

Future work will apply the LITE paradigm to other ReID-based trackers, making them faster and more practical while maintaining accuracy. Additionally, a deeper investigation into the usefulness of ReID components in various scenarios and evaluating LITE applied trackers' effectiveness on edge devices is planned.

\bibliographystyle{splncs04}
\bibliography{main}

\begin{thebibliography}{10}
\providecommand{\url}[1]{\texttt{#1}}
\providecommand{\urlprefix}{URL }
\providecommand{\doi}[1]{https://doi.org/#1}

\bibitem{aharon2022bot}
Aharon, N., Orfaig, R., Bobrovsky, B.Z.: Bot-sort: Robust associations
  multi-pedestrian tracking. arXiv preprint arXiv:2206.14651  (2022)

\bibitem{alikhanov2023online}
Alikhanov, J., Kim, H.: Online action detection in surveillance scenarios: A
  comprehensive review and comparative study of state-of-the-art multi-object
  tracking methods. IEEE Access  (2023)

\bibitem{bewley2016simple}
Bewley, A., Ge, Z., Ott, L., Ramos, F., Upcroft, B.: Simple online and realtime
  tracking. In: 2016 IEEE international conference on image processing (ICIP).
  pp. 3464--3468. IEEE (2016)

\bibitem{brostrom2023boxmot}
Brostrom, M.: Boxmot: pluggable sota tracking modules for segmentation, object
  detection and pose estimation models (2023),
  \url{'https://github.com/mikel-brostrom/yolo\_tracking'}, gitHub repository

\bibitem{cao2023observation}
Cao, J., Pang, J., Weng, X., Khirodkar, R., Kitani, K.: Observation-centric
  sort: Rethinking sort for robust multi-object tracking. In: Proceedings of
  the IEEE/CVF Conference on Computer Vision and Pattern Recognition. pp.
  9686--9696 (2023)

\bibitem{ciaparrone2020deep}
Ciaparrone, G., S{\'a}nchez, F.L., Tabik, S., Troiano, L., Tagliaferri, R.,
  Herrera, F.: Deep learning in video multi-object tracking: A survey.
  Neurocomputing  \textbf{381},  61--88 (2020)

\bibitem{CLIMENTPEREZ2020112847}
Climent-Pérez, P., Spinsante, S., Mihailidis, A., Florez-Revuelta, F.: A
  review on video-based active and assisted living technologies for automated
  lifelogging. Expert Systems with Applications  \textbf{139},  112847 (2020).
  \doi{https://doi.org/10.1016/j.eswa.2019.112847},
  \url{https://www.sciencedirect.com/science/article/pii/S0957417419305494}

\bibitem{meva_challenge}
Corona, K., Osterdahl, A., Collins, R., et~al.: Meva: A large-scale multiview,
  multimodal video dataset for activity detection. In: Proceedings of the
  IEEE/CVF Winter Conference on Applications of Computer Vision. pp. 1060--1068
  (2021)

\bibitem{MOT20}
Dendorfer, P., Rezatofighi, H., Milan, A., Shi, Q., Cremers, D., Reid, I.,
  Roth, S., Schindler, K., Leal-Taix{\'e}, L.: Mot20: A benchmark for
  multi-object tracking in crowded scenes. arXiv preprint arXiv:2003.09003
  (2020)

\bibitem{du2023strongsort}
Du, Y., Zhao, Z., Song, Y., Zhao, Y., Su, F., Gong, T., Meng, H.: Strongsort:
  Make deepsort great again. IEEE Transactions on Multimedia  (2023)

\bibitem{du2023strongsort_code}
Du, Y., Zhao, Z., Song, Y., Zhao, Y., Su, F., Gong, T., Meng, H.: Strongsort:
  Make deepsort great again (code repository).
  \url{https://github.com/dyhBUPT/StrongSORT} (2023), accessed: 2024-07-20

\bibitem{geiger2013vision}
Geiger, A., Lenz, P., Stiller, C., Urtasun, R.: Vision meets robotics: The
  kitti dataset. The International Journal of Robotics Research
  \textbf{32}(11),  1231--1237 (2013)

\bibitem{yolov8}
Jocher, G.: Yolov8. \url{https://github.com/ultralytics/ultralytics} (2023),
  accessed: November 16, 2023

\bibitem{lin2017feature}
Lin, T.Y., Doll{\'a}r, P., Girshick, R., He, K., Hariharan, B., Belongie, S.:
  Feature pyramid networks for object detection. In: Proceedings of the IEEE
  conference on computer vision and pattern recognition. pp. 2117--2125 (2017)

\bibitem{luiten2020IJCV}
Luiten, J., Osep, A., Dendorfer, P., Torr, P., Geiger, A., Leal-Taix{\'e}, L.,
  Leibe, B.: Hota: A higher order metric for evaluating multi-object tracking.
  International Journal of Computer Vision pp. 1--31 (2020)

\bibitem{luo2021multiple}
Luo, W., Xing, J., Milan, A., Zhang, X., Liu, W., Kim, T.K.: Multiple object
  tracking: A literature review. Artificial Intelligence  \textbf{293},  103448
  (2021)

\bibitem{maggiolino2023deep}
Maggiolino, G., Ahmad, A., Cao, J., Kitani, K.: Deep oc-sort: Multi-pedestrian
  tracking by adaptive re-identification. arXiv preprint arXiv:2302.11813
  (2023)

\bibitem{MOT17}
Milan, A., Leal-Taix{\'e}, L., Reid, I., Roth, S., Schindler, K.: Mot17: An
  evaluation benchmark for multi-object tracking. arXiv preprint
  arXiv:1705.02953  (2017)

\bibitem{sathyanarayana2018vision}
Sathyanarayana, S., Satzoda, R.K., Sathyanarayana, S., Thambipillai, S.:
  Vision-based patient monitoring: a comprehensive review of algorithms and
  technologies. Journal of Ambient Intelligence and Humanized Computing
  \textbf{9},  225--251 (2018)

\bibitem{personpath22}
Shuai, B., Bergamo, A., Buechler, U., Berneshawi, A., Boden, A., Tighe, J.:
  Large scale real-world multi person tracking. In: European Conference on
  Computer Vision. Springer (2022)

\bibitem{ultralytics2023multiobject}
Ultralytics: Multi-object tracking with ultralytics yolo (2023),
  \url{https://github.com/ultralytics/ultralytics/tree/main/ultralytics/trackers},
  gitHub repository

\bibitem{wojke2018deep}
Wojke, N., Bewley, A.: Deep cosine metric learning for person
  re-identification. In: 2018 IEEE winter conference on applications of
  computer vision (WACV). pp. 748--756. IEEE (2018)

\bibitem{Wojke2017simple}
Wojke, N., Bewley, A., Paulus, D.: Simple online and realtime tracking with a
  deep association metric. In: 2017 IEEE International Conference on Image
  Processing (ICIP). pp. 3645--3649. IEEE (2017).
  \doi{10.1109/ICIP.2017.8296962}

\bibitem{yu2022argus++}
Yu, L., Qian, Y., Liu, W., Hauptmann, A.G.: Argus++: Robust real-time activity
  detection for unconstrained video streams with overlapping cube proposals.
  In: Proceedings of the IEEE/CVF Winter Conference on Applications of Computer
  Vision. pp. 112--121 (2022)

\bibitem{zeng2022motr}
Zeng, F., Dong, B., Zhang, Y., Wang, T., Zhang, X., Wei, Y.: Motr: End-to-end
  multiple-object tracking with transformer. In: European Conference on
  Computer Vision. pp. 659--675. Springer (2022)

\bibitem{zhang2022bytetrack}
Zhang, Y., Sun, P., Jiang, Y., Yu, D., Weng, F., Yuan, Z., Luo, P., Liu, W.,
  Wang, X.: Bytetrack: Multi-object tracking by associating every detection
  box. In: European Conference on Computer Vision. pp. 1--21. Springer (2022)

\bibitem{zhang2021fairmot}
Zhang, Y., Wang, C., Wang, X., Zeng, W., Liu, W.: Fairmot: On the fairness of
  detection and re-identification in multiple object tracking. International
  journal of computer vision  \textbf{129},  3069--3087 (2021)

\end{thebibliography}

\end{document}